\documentclass[final, 5p, times, twocolumn]{elsarticle}
\makeatletter
\usepackage{geometry}
\usepackage{booktabs}
\usepackage{multirow}
\usepackage{caption}%
\usepackage{graphicx}
\usepackage{amssymb}
\usepackage[table]{xcolor}
\usepackage{multicol}
\usepackage{xspace}
\usepackage{enumerate}
\usepackage{wrapfig}

\makeatletter
\def\bs{\expandafter\@gobble\string\\}
\def\lb{\expandafter\@gobble\string\{}
\def\rb{\expandafter\@gobble\string\}}
\def\@pdfauthor{C.V.Radhakrishnan}
\def\@pdftitle{elsarticle.cls -- A documentation}
\def\@pdfsubject{Document formatting with elsarticle.cls}
\def\@pdfkeywords{LaTeX, Elsevier Ltd, document class}

\DeclareRobustCommand{\LaTeX}{L\kern-.26em%
        {\sbox\z@ T%
         \vbox to\ht\z@{\hbox{\check@mathfonts
           \fontsize\sf@size\z@
           \math@fontsfalse\selectfont
          A\,}%
         \vss}%
        }%
     \kern-.15em%
    \TeX}
\makeatother

\usepackage{lineno}
\usepackage{algorithm}
\usepackage{booktabs}
\usepackage{verbatim}
\usepackage{array}
\usepackage{subfigure}
\usepackage[utf8]{inputenc}
\usepackage{pifont}
\usepackage{float}
\usepackage{placeins}
\usepackage{algpseudocode}
\newcommand{\xmark}{\ding{55}}
\usepackage{amsmath}
\usepackage{tabularx}
\newcolumntype{L}{>{\raggedright\arraybackslash\hangindent=1em\hangafter=1}X}
\usepackage[colorlinks,backref=true]{hyperref}
\usepackage[font=small,labelfont=bf,tableposition=top]{caption}
\AtBeginDocument{\hypersetup{citecolor=cyan,linkcolor=cyan,urlcolor=cyan,pdfhighlight=N,}}
\newcolumntype{L}{>{\centering\arraybackslash}m{2cm}}
\newcolumntype{K}{>{\centering\arraybackslash}m{0.5cm}}

\begin{document}
\begin{frontmatter}

\title{Outlier-Oriented Poisoning Attack: A Grey-box Approach to Disturb Decision Boundaries by Perturbing Outliers in Multiclass Learning}

\author{Anum Paracha} 
\author{Junaid Arshad} 
\author{Mohamed Ben Farah}
\author{Khalid Ismail}
\address{College of Computing, Birmingham City University, UK}


\begin{abstract}
Poisoning attacks are a primary threat to machine learning models, aiming to compromise their performance and reliability by manipulating training datasets. This paper introduces a novel attack - Outlier-Oriented Poisoning (OOP) attack, which manipulates labels of most distanced samples from the decision boundaries. The paper also investigates the adverse impact of such attacks on different machine learning algorithms within a multiclass classification scenario, analyzing their variance and correlation between different poisoning levels and performance degradation. To ascertain the severity of the OOP attack for different degrees (5\% - 25\%) of poisoning, we analyzed variance, accuracy, precision, recall, f1-score, and false positive rate for chosen ML models. Benchmarking our OOP attack, we have analyzed key characteristics of multiclass machine learning algorithms and their sensitivity to poisoning attacks. Our experimentation used three publicly available datasets: IRIS, MNIST, and ISIC. Our analysis shows that KNN and GNB are the most affected algorithms with a decrease in accuracy of 22.81\% and 56.07\% while increasing false positive rate to 17.14\% and 40.45\% for IRIS dataset with 15\% poisoning. Further, Decision Trees and Random Forest are the most resilient algorithms with the least accuracy disruption of 12.28\% and 17.52\% with 15\% poisoning of the IRIS dataset. 
We have also analyzed the correlation between number of dataset classes and the performance degradation of models. Our analysis highlighted that number of classes are inversely proportional to the performance degradation, specifically the decrease in accuracy of the models, which is normalized with increasing number of classes. Further, our analysis identified that imbalanced dataset distribution can aggravate the impact of poisoning for machine learning models. 
\end{abstract}

\end{frontmatter}
\section{Introduction}
The widespread use of machine learning in diverse fields such as cyber security \cite{han2023credential}, healthcare \cite{moradi2023recent}, and autonomous vehicles \cite{zeng2024convergence} make it an attractive target for adversaries. Machine learning models are susceptible to various types of adversarial attacks, typically classified as poisoning \cite{chillara2024deceiving}, evasion \cite{bostani2024evadedroid}, backdoor \cite{yang2024stealthy}, inversion \cite{nguyen2024label} and inference \cite{liu2023gradient}, \cite{ali2023adversarial} attacks. Some potential attacks are demonstrated in \cite{li2022query, 10.1145/3576915.3616661, jha2024label, manna2024trimpa, yu2024chronic}. Of these, poisoning attacks are one of the most common in literature, aiming to manipulate training datasets to corrupt machine learning models. Poisoning attacks are either aimed at the overall performance degradation ( \textit{availability attacks}) or targeted to mislead on specific instances (\textit{integrity attacks}). As a reasonably large amount of data is required for model training, it enables adversaries to manipulate and poison datasets that may be difficult to cleanse. Prior research on poisoning availability and integrity attacks against machine learning is mostly focused on deep learning and binary classification, however, exploring poisoning effects on multiclass classifiers is limited. \\
Practical interpretation of poisoning availability attacks and their impact on performance degradation on machine learning models have been studied extensively in literature such as \cite{10.5555/3042573.3042761, wang2024efficient, lorenz2023certifiers, liu2023image, jonnalagadda2024modelling, di2022hidden}. An effective approach of creating poisoned data points with a generative adversarial network (GAN) in the same underlying distribution as the irreducible noise of the classifier is presented in \cite{zhao2022clpa}. C. Zhou et al. \cite{zhou2024object} developed an object-attentional adversarial attack and generated a unique dataset with adversarial images. Another stealthy data poisoning attack is developed in \cite{schneider2023dual} through which adversarial samples exhibit the properties of desired samples. Similarly, \cite{koh2022stronger} presented an approach to craft data points so they are placed near one another to bypass data sanitization by evading anomaly detection in the spam detection classifier. Further, X. Zhang et al. \cite{zhang2020online} studied the adverse impact of data poisoning in online settings which highlighted its effectiveness against reinforcement learning. In research studies \cite{chen2021deeppoison, zhang2023clean, zhong2020backdoor}, features perturbation methods are used to poison training of the machine learning. From the literature, roughly three approaches are used for data poisoning attacks. Firstly, label poisoning \cite{liu2022adversarial, shahid2022label, aryal2022analysis} which typically follows the label-flipping method for poisoning training datasets. Secondly, clean-label poisoning \cite{zhu2019transferable, aghakhani2021bullseye} is formally crafted by solving one or more optimization algorithms such as bi-level optimization \cite{9448491, russo2021poisoning}, or gradient-descent \cite{10.1145/3373376.3378462} to craft poisoned data points and inject into the machine learning model. And thirdly, the existing dataset is manipulated with feature perturbation. \\
However, poisoning attacks against multiclass machine learning are explored on a limited scale. MetaPoison \cite{huang2020metapoison} solves the bi-level optimization with meta-learning to craft poison against neural networks. It is practically implemented on Google Cloud AutoML API and extended to be experimented on multiclass neural networks. Sub-population data poisoning \cite{10.1145/3460120.3485368} populates a perturbed cluster into the target. Its efficacy is highlighted with a variety of neural networks with multiple datasets. Another research study \cite{munoz2017towards} proposed a back-gradient based poisoning attack and extended their experimentation from binary classification to multiclass classification. Experimentation of this research study is focused on the poisoning availability of targeted subclass in neural networks. Also, existing research is limited to the analysis of performance degradation only, considering the accuracy of the model, mostly considering poisoning neural networks only. \\

In view of the limitations highlighted above, we have explicitly identified the need to analyze the behavior of individual algorithms against poisoning attacks in multiclass settings. We have extended the experimentation of multiclass poisoning attack and proposed a novel attack method, with outliers poisoning, to poison multiclass supervised machine learning. Mentioning multiclass poisoning from literature that is mostly limited to the poisoning of neural networks, we leverage our research to study poisoning on six supervised machine learning algorithms including Support Vector Machines (SVM), Decision Tree (DT), Random Forest (RF), K-Nearest Neighbors (KNN), Gaussian Naive Bayes (GNB) and Neural networks Multi-layer Perceptron(MLP). Selecting these algorithms allows us to cover the complete baseline of machine learning classification methods. With our analysis, we have identified important parameters of each algorithm that are sensitive to poisoning attacks, answering how the models are getting misclassified and identifying optimal poisoning rates for each algorithm. We have provided an in-depth investigation of the performance degradation of multiclass models, quantifying the accuracy variance of the models. Our research contributions are given as follows:

\begin{itemize}

\item A thorough behavioral analysis of multiclass classifiers is conducted analyzing the correlation between different poisoning levels and performance degradation of classifiers. Using a range of poisoning levels $\Delta L=5\%-25\%$, we highlight the optimal rate of poisoning that can mislead classification in an obscure manner.

\item We analyze the behavior of individual machine learning algorithms in multiclass settings and the impact of the noisy dataset on our poisoning attack. We successively study the impact of non-uniformly structured features and imbalanced dataset distribution. We further explained different behaviors of poisoned datasets where it work as a catalyst for data poisoning, leading to an impractical model disruption. 

\item We have developed a novel label poisoning attack to introduce misclassification in multiclass machine learning. Our attack is formulated based on the label perturbation of the most distanced data points from the decision boundaries of the multiclass classifier where the data points distances are calculated by training surrogate models.

\item By implementing our attack, we analyze the key factors for each algorithm that are affected by data poisoning. For our analysis, we have implemented our attack on six machine learning algorithms with IRIS, MNIST, and ISIC datasets. Our research findings serve as a baseline to strengthen mitigation against data poisoning in multiclass models.
 \end{itemize}

\section{Existing Multiclass Machine Learning Poisoning}
\subsection{Existing Attacks to Poison Multiclass Models}
Existing literature highlights a significant count of poisoning attacks that harm the integrity and availability of machine learning models. Such as B. Zhao et al. \cite{zhao2022towards} proposed a class-oriented poisoning attack to introduce misclassification for a targeted dataset class. Similarly, N. Carlini et al. \cite{carlini2021poisoning} highlighted a security threat of poisoning and backdoor attacks against multiclass machine learning with only 0.0001\% of data poisoning. They have introduced misclassification in the targeted model with training-time overfitting to increase error at testing. I. Alarab et al. \cite{alarab2023uncertainty} have developed a Monte-Carlo based poisoning attack against deep learning multiclass models to analyze their classification uncertainties. Whereas, V. Pantelakis et al. \cite{pantelakis2023adversarial} have evaluated the performance disruption of IoT-based multiclass models against JSMA, FGSM, and DeepFool attacks with which effectiveness of these attacks are highlighted to poison multiclass models. Also, some other prominent poisoning attacks are \cite{suya2021model, mayerhofer2022poisoning, saha2020hidden}. 
\begin{table*}[h]
\small
    \centering
    \caption{Analyzing existing studies against our behavioral analysis with OOP attack}
    \label{Analyzing existing studies against our behavioral analysis with OOP attack}
    \begin{tabular}
    {p{8em} p{10em} p{8em} p{8em} p{9em} p{6em}}
\toprule
Research paper & ML model & Dataset & Effective poisoning level & \multicolumn{2}{ c}{Model degradation and variance at:}\\
\cline{5-6}
&&&&Various poisoning levels & Various classes\\
 \midrule
B. Zhao et al. \cite{zhao2022towards} & LeNet-5, Vgg-9, ResNet-50 & MNIST, CIFAR-10, ImageNet & \xmark & \xmark & \xmark\\
N. Carlini et al. \cite{carlini2021poisoning} & ResNet-50, Transformer language model & Conceptual Captions & \xmark & \xmark & \xmark\\
I. Alarab et al. \cite{alarab2023uncertainty} & LEConv, CNN & Cora, MNIST & \xmark & \xmark & \xmark\\
V. Pantelakis et al. \cite{pantelakis2023adversarial} & DT, RF, KNN, MLP & IoTID20 & \xmark & \xmark & \xmark\\
    \bottomrule 
    \end{tabular}
\end{table*}
Existing studies have experimented with complex deep learning and machine learning models, given in Table \ref{Analyzing existing studies against our behavioral analysis with OOP attack}. Whereas, it is important to understand the behavior of the underlying baseline models and their sensitivity against poisoning attacks. This investigation helps us better mitigate poisoning not only focusing on their performance but underlying classification mechanisms. Considering the above highlighted attack, our work has focused on manipulating outliers to disrupt the features spaces of the multiclass models, discussed in Section \ref{outlier-oriented Poisoning (OOP) Attack Method}. In this work, we have shown the efficacy and effectiveness of our attack on six machine learning algorithms at various poisoning levels. 
\subsection{Existing Security Techniques against Multiclass Model Poisoning}
Limited techniques are provided in literature to cleanse datasets to mitigate poisoning effects against multiclass machine learning models. A. McCarthy et al. \cite{mccarthy2023defending} proposed a hierarchical learning mechanism to secure network traffic attack classification model. K. M. Hossain et al. \cite{hossain2024advancing} developed a solution to detect backdoor poison in deep neural networks by extracting, relabeling, and classifying features with a tensor decomposition method. They have experimented their mitigation solution with MNIST, CIFAR-10, and TrojAI datasets. Curie \cite{laishram2016curie} is the method proposed to mitigate poisoning attacks against SVM. This method has introduced an additional feature dimension to map labels with features that help segregate the poisoned data points with flipped labels from the normal data points. S. Melacci et al. \cite{melacci2021domain} have experimented the effectiveness of incorporating domain knowledge, into the neural networks, in detecting adversarial data points added in the model training. They have experimented their solution on neural networks with CIFAR-100, ANIMALS, and PASCAL-Part datasets. Highlighting the above solutions which explained the efficacy of their proposed poisoning methods, we have explicitly mentioned the need for a deep behavioral analysis of multiclass classifiers under poisoning attacks to identify key characteristics of machine learning algorithms affected by poison and reveal relations between those characteristics and the injected poison. Further to achieve this, we have discussed our threat model and attack formulation in Section \ref{Formal Notations of multiclass Poisoning}.
\section{Formal Notations of Multiclass Poisoning}
\label{Formal Notations of multiclass Poisoning}
\subsection{Threat Model}
\label{Threat Model}
We have taken a practical approach when designing our attack strategy, aligning our attack settings with the assumption that the adversary $A_{dv}$ does not know the underlying settings of the targeted model $M_{c}$ and dataset distribution $dist(D_{c})$. Three different datasets—IRIS, MNIST, and ISIC—each with three, ten, and four classes respectively, were used to assess our methodology. We have formulated our attack in end-to-end settings of poisoned training, of benign models, where only datasets are known to the adversary. Surrogate models $M_{s}$ are developed and trained for crafting poisoned data points $X_{c}'$ with perturbed class labels $l_{c}'$ at different poisoning levels $\Delta{L}$. Poisoning levels are set between 5\%-25\% inclusively, at the scale of 5. Our OOP attack initiates multiclass poisoning by calculating the maximum distanced data points to be contaminated and changing their classes ${l_{c}'}$. With this outliers perturbation, we manipulate benign features spaces by misplacing outliers in those features spaces. The goals of our OOP attack are to assess the performance degradation of individual algorithms and analyze the behavior of multiclass models, under the OOP attack. For the dataset manipulation with our OOP attack, let $X_{c} \in D_{c}$ be the distanced data point that is infected by manipulating its label $l_{c}$ to increase the loss of the model $\mathcal{L}$ with change $\gamma$ in multiclass decision boundaries $b_{c}$ as:
\begin{equation}
\mathcal{L}(M_{c}, D_{c}') = \gamma = \Delta b_{c}(\mathcal{T}(M_{c},D_{c}'))\\
\label{Eq:LossFunction}
\end{equation}
where $\mathcal{T}$ is the model training process.
\subsection{Outlier-Oriented Poisoning (OOP) Attack Settings}
\label{outlier-oriented Poisoning (OOP) Attack Settings}
We have developed our attack for analyzing the classification disruption of individual algorithms in multiclass classification settings. We are developing surrogate models to poison datasets because based on the assumptions of our threat model, given Section \ref{Threat Model}, we do not know about configurations of victim models. Firstly, we have initialized surrogate models with each algorithm and trained them with the targeted datasets. Secondly, the distances of each data point are calculated from the decision boundary for each class to manipulate those far from the decision boundaries following our attack settings with Definitions \ref{Definition 1 (multiclass Poisoned Training)}, \ref{Definition 2 (multiclass Model Disruption)} and \ref{Definition 3 (multiclass Performance Analysis)} where $D_{c}$ is the dataset with the \textit{n} number of classes and $M_{c}$ is the clean model and $M_{c}'$ is the poisoned model. 
\subsubsection{Definition 1 (Multiclass Poisoned Training)}
\label{Definition 1 (multiclass Poisoned Training)} Considering $\mathcal{T}$ is the model training process with poisoned dataset $D_{c}'$ and $\mathcal{P_{M_{c}}}$ is the function of performance measure, the objective function of our attack method is given in Equation \ref{Eq:ObjectiveFunction} whereas, $\theta$ is the measure of distance of data points from the decision boundaries. Mathematical notation of $\theta$ is given in Equation \ref{Eq:DistanceVector}.
\begin{equation}
\begin{aligned}
\arg \min \: \mathcal{P_{M_{c}}}(M_{c}(X_{c}',l_{c}'); \theta))\\
\end{aligned}
\label{Eq:ObjectiveFunction}
\end{equation}
\begin{equation}
\begin{aligned}
\textbf{s.t.} \: \theta = \arg \max \: \vec{d}(b_{c}, X_{c})\\
\end{aligned}
\label{Eq:DistanceVector}
\end{equation}
Also, $D'_{c}$ is the poisoned dataset manipulated at various poisoning levels $\Delta L$ where the notation of dataset poisoning is given in Equation \ref{Eq:DatasetPoisoning}.
\begin{equation}
\begin{aligned}
D'_{c} = \sum_{i=1}^{n \rightarrow \Delta L}{f(D_{c_i}(X_{c_i}, l_{c_i}), \Delta L)}\\
\textbf{where;}\;
X_{l_{c}} \neq X'_{l_{c}}\\
\end{aligned}
\label{Eq:DatasetPoisoning}
\end{equation}
where, \textit{f} is the function of manipulating labels, $X_{c}$ is the clean data point, $X'_{c}$ is the poisoned data point and $l_{c}$ is the label. 

\subsubsection{Definition 2 (Multiclass Model Disruption)}
\label{Definition 2 (multiclass Model Disruption)} Let poison levels $\Delta L$ = [5\%,10\%,15\%,20\%,25\%] manipulate datasets to mislead models at test-time by disturbing class-level decision boundaries $b_{c}$ with notation given in Alg 
\ref{Distanced-based Poisoned Training}. Let \textit{f} is the function to poison dataset $D_c$ at $\Delta L$ poison level. $M'_{c}$ is the poisoned model trained with a dataset having manipulated data points $X_{c}'$ as given in Eq \ref{Poisoned Model}.
\begin{equation}
\begin{aligned}
M'_{{c}} = \mathcal{T}(M_{c}, D_{c}')\\
\textbf{where;}\; 
D_{c}'= f(D_{c}(X_{c}, l_{c}), \Delta L)
\end{aligned}
\label{Poisoned Model}
\end{equation}
which allows us to analyze the model behavior and change in decision boundaries as given in Eq \ref{Model Disruption}.
\begin{equation}
Mod_{Dis} = \Delta{b}(M_{c}')
\label{Model Disruption}
\end{equation}
where $M'_{c}$ is the poisoned model developed for algorithms [SVM, DF, RF, KNN, GNB, MLP] and \textit{$\Delta b$} is the change in decision boundaries.
\subsubsection{Definition 3 (multiclass Performance Analysis)}
\label{Definition 3 (multiclass Performance Analysis)} To conduct a statistical analysis of the performance degradation of multiclass models and the variance in test-time classification across different poisoning levels, we define the correct classification rate given by Eq \ref{Performance degradation}.  
\begin{equation}
\begin{aligned}
C_{CR} = \frac{\sum_{i=1}^{n}{f(N_{{c}},C_{{c}_{i}}(M_{{c}_{i}}'(D_{t}(X_{{t}_{i}}))))}}{\sum_{i=0}^{n=1}{N_{{c}}}}\\
and \: f(N_{c}, C(D_{t}(X_{t}))) = \begin{cases}
    true & \text{if $X_{t} \in Class \; c$}\\
    false & \text{otherwise}
\end{cases}
\end{aligned}
\label{Performance degradation}
\end{equation}
where, \textit{f} is the function of classification, $X_{t}$ is the data point from the validation dataset $D_{t}$, $N_{c}$ is the total number of data points in Class c, $C(.)$ is the class estimator and $C_{CR}$ is the unpoisoned classification rate.
\section{Outlier-Oriented Poisoning (OOP) Attack Method}
\label{outlier-oriented Poisoning (OOP) Attack Method}
Instinctively, we are poisoning the training dataset to disrupt machine learning performance at validation. The proposed outlier-oriented poisoning attack algorithm is given in Alg \ref{Distanced-based Poisoned Training}, With the OOP attack, we are focusing on manipulating labels of most distanced data points from the class boundaries to shift the classification predictions.
This approach follows the threat model outlined in Section \ref{Threat Model}. Based on our threat model, we don't have any configurations details of the targeted model and have developed surrogate models to calculate data points distances. The surrogate model development algorithm is presented in Alg. \ref{Surrogate Model Development} for training surrogate models. Furthermore, we have enhanced our attack strategy by distinctively calculating decision boundaries for the considered machine learning algorithms. The algorithm to calculate decision boundaries is described in Alg. \ref{Calculating Distances from Decision Boundaries}, following the attack settings given in Section \ref{outlier-oriented Poisoning (OOP) Attack Settings}.
\begin{algorithm*}[h]
\caption{OOP Poisoned Model Generation}
\label{Distanced-based Poisoned Training}
\begin{algorithmic}
\State \textbf{Datasets}: IRIS, MNIST, ISIC, ChestX-ray-14 datasets
\State \textbf{Inputs}: Training Dataset $D_c$, Poison level $\Delta L$
\State \textbf{Outputs}: Poisoned Model $M_c'$
\State \textbf{Initialize}: $D_c \gets$ Training dataset
\State $\Delta L \gets$ Poisoning level $\in [0\%, 5\%, 10\%, 15\%, 20\%, 25\%]$
\State $M_{conf} \gets [SVM, DT, RF, GNB, KNN, MLP]$
\State $D_c' \gets$ Poisoned dataset = $[]$
\State $D_{dist} \gets$ subset of Training dataset 
\While{$len(D_c') \leq \Delta L$}
    \State Set index i = max($D_{dist}$)
    \State Set data point $d_c$ = $D_c[i]$
        \If{$d_c$ not in $D_c'$}
            \State Set $l_c$ = Class($d_c$)
            \State Update $l_c$ = $l_x$; where $x \neq c$
            \State Update Class($d_c'$) = $l_x$
        \EndIf
        \State $D_c' \gets d_c'$
        \State Set $D_{dist}[i] = 0$
\EndWhile\\
$D_{ctrain}'$ = split($D_c'$, 0.75)\\
$M_c'$ = train($M_{conf}$, $D_{ctrain}'$)\\
\textbf{return} $M_c'$
\end{algorithmic}
\end{algorithm*}

\begin{algorithm*}[h]
\caption{Surrogate Model Development}
\label{Surrogate Model Development}
\begin{algorithmic}
\State \textbf{Datasets}: IRIS, MNIST, ISIC, ChestX-ray-14 datasets
\State \textbf{Inputs}: Training Dataset $D_c$, Model Configuration $M_{conf}$
\State \textbf{Outputs}: Surrogate Trained Model $M_{surr}$
\State \textbf{Initialize}: $D_c \gets$ Training dataset
\State $M_{conf} \gets$ [
Support Vector Machines (SVM) = Config(kernel='poly', degree of polynomial function=3, regularization parameter=3),\\
Decision Tree(DT) = Config(criterion='gini', splitter='best')\\
Random Forest(RF) = Config(n\_estimators=3, criterion='gini')\\
K-Nearest Neighbours (kNN) = Config(n\_neighbors=5, weights='uniform')\\
Gaussian Naive Bayes (GNB) = Config(var\_smoothing=$1*10^{-9}$)\\
Multi-layer Perceptron (MLP) = Config(activation='relu', solver='adam')]\\

\For{config in $M_{conf}$}
    \State $M_{surr}(config) = initialize(M_{surr}, config)$
    \State $M_{surr}(config) = training(M_{surr}(config), D_{c})$
\EndFor\\
\textbf{return} $M_{surr}(config)$
\end{algorithmic}
\end{algorithm*}

\begin{algorithm*}
\caption{Calculating Distances from Decision Boundaries}
\label{Calculating Distances from Decision Boundaries}
\begin{algorithmic}
\State \textbf{Inputs}: Surrogate Models $M_{surr}$, Training Dataset $D_{c}$
\State \textbf{Outputs}: Calculated distances of Models $dist_{M}$
\State \textbf{Initialize}: $dist_{M} = [dist_{SVM}, dist_{DT}, dist_{RF}, dist_{GNB}, dist_{KNN}, dist_{MLP}]$
\State $d_p \gets$ Model data points
\State $M_{surr} = [M_{SVM}, M_{DT}, M_{RF}, M_{GNB}, M_{KNN}, M_{MLP}]$
\If{$M_{surr} == M_{SVM}$}
    \For{$d_p \in M_{SVM}$}
        \State $dist_{SVM}[d_p] \gets decision_function(d_p, M_{SVM})$
    \EndFor
    $dist_{M}[SVM] = d_{p}$
\EndIf
\If{$M_{surr} == M_{DT}$}
    \State $Clf_{tree} = M_{DT}.tree\_$
    \For{$d_p \in D_c$}
        \State $dist[d_p] \gets calculate\_depth(d_p, Clf_{tree})$
    \EndFor
    $dist[DT] = d_{p}$
\EndIf
\If{$M_{surr} == M_{RF}$}
    \For {$clf_x \in M_{RF}$}
    \State $Clf_{tree} = clf_x.tree\_$
    \For{$d_p \in D_c$}
        \State $dist[d_p]_x=calculate\_depth(d_p, Clf_{tree})$
    \EndFor
    \EndFor
    \State $dist_{M}[RF] = avg(dist[d_p]_{x}, dist[d_p]_{x+1}, ... dist[d_p]_{x+n})$
\EndIf
\If{$M_{surr} == M_{KNN}$}
    \For{$d_p \in D_c$}
        \State $dist(d_p)_{neighbors} = M_{KNN}.kneighbors\_$
        \State $dist_{M}[KNN] \gets \arg \max (distance(d_p)_{neighbors})$
    \EndFor
\EndIf
\If{$M_{surr} == M_{GNB}$}
    \State $D_{c_a}, D_{c_b} = split(D_c, 2)$
    \For{$i \in [D_{c_a}, D_{c_b}]$}
        \State $j = -i+1$
        \For{$d_p \in D_c[i]$}
            \State $Class(d_p)= predict_probability(D_c[j], M_{GNB})$
            \State $log_likelihood \gets log(Class(d_p))$
            \State $distance(d_p) \gets distance(\arg \max(Class(d_p), axis=1))$
        \EndFor
    \EndFor
    $dist_{M}[GNB] = distance(d_p)$
\EndIf
\If{$M_{surr} == M_{MLP}$}
    \For{$d_p \in M_{MLP}$}
        \State $dist_{M}[MLP] \gets decision_function(d_p, M_{MLP})$
    \EndFor
\EndIf\\
\textbf{return} $dist_{M}$
\end{algorithmic}
\end{algorithm*}

\subsection{Evaluation Metrics}
\label{Evaluation Metrics}
For the performance evaluation and analysis of the impact of poisoning availability attacks on multiclass supervised machine learning algorithms, we have evaluated our poisoned models by analyzing how many outliers successfully intruded themselves in wrong classes which is the \textbf{False Positive Rate(FPR)} of our poisoned model. However, where our poisoned outliers remain disjointed in the incorrect classification classes and model availability is intact is the \textbf{Accuracy(Acc)}, and where the outliers are unsuccessful in intruding the multiclass decision boundaries is the \textbf{Precision} of the model against OOP attack. \textbf{Recall} in our evaluation is the quantification where a model can segregate dataset classes and keep the decision boundaries intact. And \textbf{Variance(Var)} in the change in model behavior with the change in the values of its parameters with the discrepancy in the dataset. Considering $f(N_{c}, C(X_{t}))$ is the classification function as given in Eq \ref{Eq:ClassificationFunction}, the evaluation metrics are mentioned in Eq \ref{Eq:fpr}, \ref{Eq:accuracy}, \ref{Eq:precision}, \ref{Eq:recall} and \ref{Eq:variance}.
\begin{equation}
f(N_{c}, C(X_{t})) = \begin{cases}
    true & \text{if $X_{t} \in Class \; c$}\\
    false & \text{otherwise}
\end{cases}
\label{Eq:ClassificationFunction}
\end{equation}

\begin{equation}\label{Eq:fpr}
\begin{aligned}
FPR = \frac{\sum_{i=0}^n f_{tr}(N_{c}, C(X_{{t}_{i}}'))}{\sum_{i=0}^n f(N_{c}, C(X_{t}')) \land \sum_{i=0}^n f(N_{c}, C(X_{t}))}\\\\
where\: f_{tr}(N_{c}, C(X_{{t}_{i}}')) \in D_{c}'\\
and\: f(N_{c}, C(X_{t}')) \in D_{c}'\\
and\:f(N_{c}, C(X_{t})) \in D_c
\end{aligned}
\end{equation}
where, $D_{c}$ is the clean dataset, $N_{c}$ is the total number of data points in Class \textit{c}, and $D_{c}'$ is the poisoned dataset with changed class labels of the farthest data points. $f_{tr}(N_{c}, C(X_{{t}_{i}}'))$ are poisoned data points with perturbed labels and classified as false positives(FP) and $f_{fs}(N_{c}, C(X_{{tr}_{i}}'))$ are false negative(FN) data points. 
\begin{equation}\label{Eq:accuracy}
Acc = \frac{\sum_{i=0}^n f_{fs}(N_{c}, C(X_{{t}_{i}}))\land \sum_{i=0}^n f_{tr}(N_{c}, C(X_{{t}_{i}}))}{(X_{c} \in D_{c}) \land (X_{c}' \in D_{c}')}
\end{equation}

\begin{equation}\label{Eq:precision}
\begin{aligned}
Prn = \frac{\sum_{i=0}^n f_{tr}(N_{c}, C(X_{{t}_{i}}))}{\sum_{i=0}^nf_{tr}(N_{c}, C(X_{{t}_{i}})) \land \sum_{i=0}^nf_{tr}(N_{c}, C(X_{{t}_{i}}'))}\\\\
where\: f_{tr}(N_{c}, C(X_{{t}_{i}}')) \in D_{c}'
\end{aligned}
\end{equation}

\begin{equation}\label{Eq:recall}
\begin{aligned}
Rcl = \frac{\sum_{i=0}^n f_{tr}(N_{c}, C(X_{{t}_{i}}))}{\sum_{i=0}^n(f_{tr}(N_{c}, C(X_{{t}_{i}}))) \land \sum_{i=0}^n(f_{fs}(N_{c}, C(X_{{t}_{i}})))}\\\\
where\: f_{fs}(N_{c}, C(X'_{{t}_{i}})) \in D_{c}'
\end{aligned}
\end{equation}

\begin{equation}\label{Eq:variance}
Variance(\sigma) = \frac{1}{N_{c}}\sum_{i=0}^n (f(N_{c}, C(X_{{t}_{i}})) - \mu(f(N_{c}, C(X_{{t}_{i}}))))^2
\end{equation}

\section{Experimentation and Ablation Study}
\subsection{Experimental Setup} We have implemented our OOP attack on six multiclass machine learning algorithms, utilizing three datasets in five different poisoning settings detailed in Section \ref{outlier-oriented Poisoning (OOP) Attack Settings}. We have developed surrogate models of all six algorithms with model configurations given in Alg \ref{Surrogate Model Development} for experimentation in a grey-box scenario. Following \cite{wongrassamee2017can}, \cite{huang2022embedding}, and \cite{zheng2023trojfair}, we have selected IRIS, MNIST, and ISIC multiclass datasets for our experimentation. Section \ref{Dataset Description} gives a dataset description and analysis. For our experiments, 75\% of the dataset has been allocated for training -including poisoned training- while the remaining 25\% cleaned dataset is used for testing the model performance. This setup allowed us to assess the impact of our outlier-oriented poisoning approach on the models using evaluation metrics outlined in Section \ref{Evaluation Metrics}. All models are developed in Python with Scikit-learn, Pandas, Numpy, and OpenCV libraries, and they were experimented on a Windows 11 Pro 64-bit machine with 4 core CPUs and 16 GB RAM. 
\subsection{Datasets Description}
\label{Dataset Description}
Outlier-oriented label poisoning is implemented on multiclass classifiers using three benchmarked datasets (IRIS, MNIST, and ISIC). All the datasets are comprised of different numbers of classes, features, sizes, and structures. Datasets characteristics are provided in Table \ref{Datasets description}. By employing our attack across datasets with differing structures, we provide a comprehensive analysis of how data poisoning influences feature correlations, class numbers, and dataset sizes within multiclass contexts. 
\begin{table}[h]
\small
    \centering
    \caption{Dataset description used for behavioral analysis with distance-based attack}
    \label{Datasets description}
    \begin{tabular}{p{2em} p{3.4em} p{4em} p{5em} p{7em}}
    \toprule
         S.No. &  Dataset & No. of features &  No. of Classes & No. of instances\\
    \midrule
         1 & IRIS & 4 & 3 & 170\\
         2 & MNIST & 784 & 10 & 70,000\\
         3 & ISIC & 20 & 4 & 603\\
         \bottomrule 
    \end{tabular}
\end{table}
The visual datasets representation with the Gaussian Mixture Model is given in Figure \ref{Datasets description}, highlighting their features correlation. 
\begin{figure*}[h]
\centering
\subfigure[\small Features correlation in IRIS dataset]
    {\includegraphics[width=7.8cm,height=5.2cm]{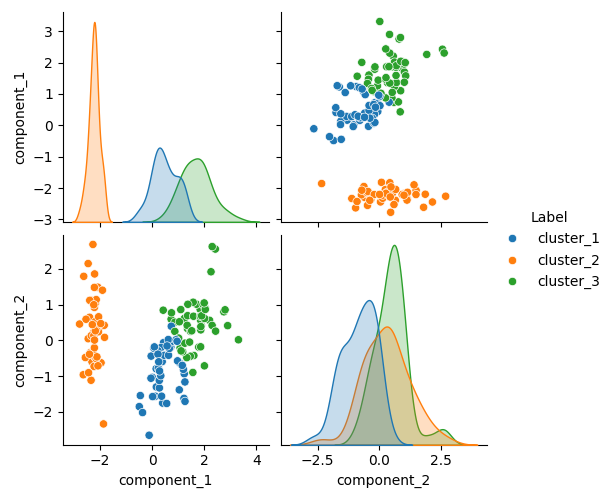}}
\subfigure[\small Features correlation in MNIST dataset]
    {\includegraphics[width=7.8cm,height=5.2cm]{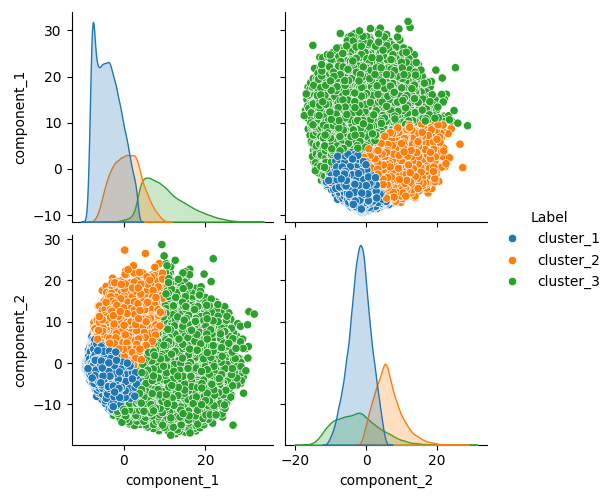}}
\subfigure[\small Features correlation in ISIC dataset]
    {\includegraphics[width=7.8cm,height=5.2cm]{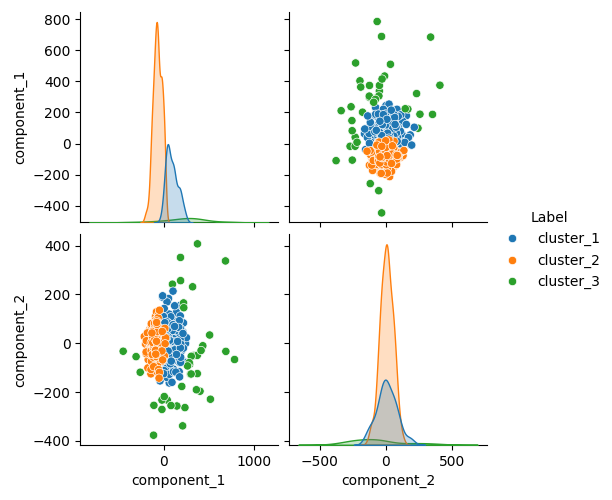}}
\caption{GMM visualization of features relationship in the dataset with PCA reduction}
\label{fig:GMM visualization of features relationship in dataset with PCA reduction}
\end{figure*}
Figure \ref{Datasets description}(a) illustrates that certain features within the IRIS dataset are strongly interdependent, whereas the complete dataset is not in a linear relation. However, MNIST is found to be a highly dense dataset with strong features relations as visualized in Figure \ref{Datasets description}(b). The ISIC dataset, shown in Figure \ref{Datasets description}(c), displays a non-linear relationship with significant outliers, indicative of substantial noise levels. Statistical correlations and feature relationships are quantified in Table \ref{Features Correlation}. Features in the MNIST dataset are highly associated with a p-value of 0.0141, highlighting direct proportionality between its features. A low statistical significance is shown in IRIS datasets with a p-value of 0.07, and the p-value of the ISIC dataset is 0.2396. In contrast, a negative Spearman correlation coefficient highlights a negative linear correlation between its features with a high noise ratio. Further analysis of the importance of features correlation and impact of dataset noise in our OOP Attack is given in Section \ref{Experimental Results and Analysis}.
\begin{table}[h]
    \centering
    \caption{Statistical correlation of features in dataset}
    \label{Features Correlation}
    \begin{tabular}{c c c c}
    \toprule
         S.No. &  Dataset & Spearman Correlation & p-value\\
    \midrule
         1 & IRIS & 0.123888 & 0.0791\\
         2 & MNIST & 0.009282 & 0.0141\\
         3 & ISIC & -0.014311 & 0.2396\\
         \bottomrule 
    \end{tabular}
    \vspace{-0.4cm}
\end{table}
\subsection{Experimental Results and Analysis}
\label{Experimental Results and Analysis}
In our study, we conducted experimental evaluations of our OOP attack on multiclass machine learning algorithms. Our objective was to analyze the behavior of multiclass models and answer questions about how the characteristics of these models are affected and identify their relationship with the poison. What are the complacent poisoning levels $\Delta L$ and effects of changing poisoned data distributions? What is the effectiveness and persistence of data poisoning with our OOP attack and its impact on model validation performance (specifically accuracy)? And quantifying and analyzing model variance $\sigma$ at test-time classification at different poisoning levels $\Delta L$.
\subsubsection{Effects of OOP Attack and Factors Affecting Multiclass Model Classification}
\begin{figure*}[!t]
\centering
\subfigure[\small Poisoning SVM with IRIS dataset]
    {\includegraphics[width=5.8cm,height=3.1cm]{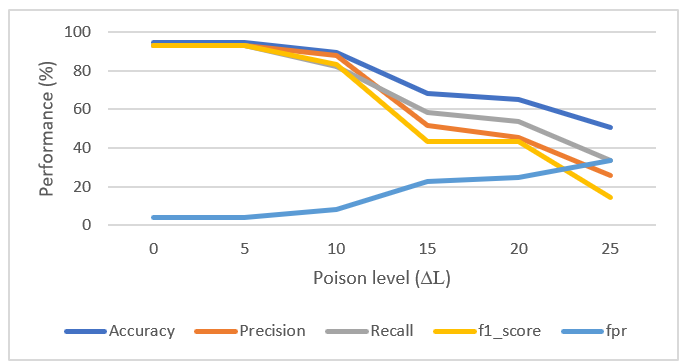}}
\subfigure[\small Poisoning SVM with MNIST dataset]
    {\includegraphics[width=5.8cm,height=3.1cm]{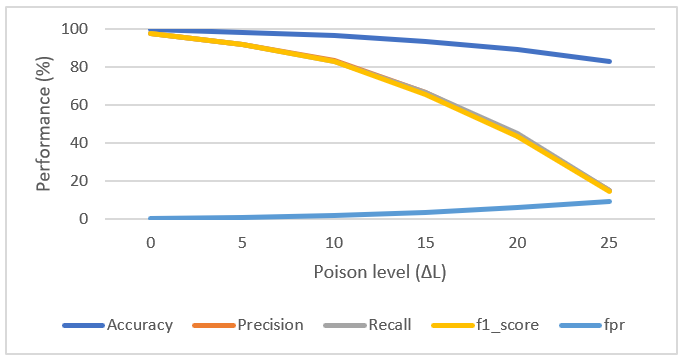}}
\subfigure[\small Poisoning SVM with ISIC dataset]
    {\includegraphics[width=5.8cm,height=3.1cm]{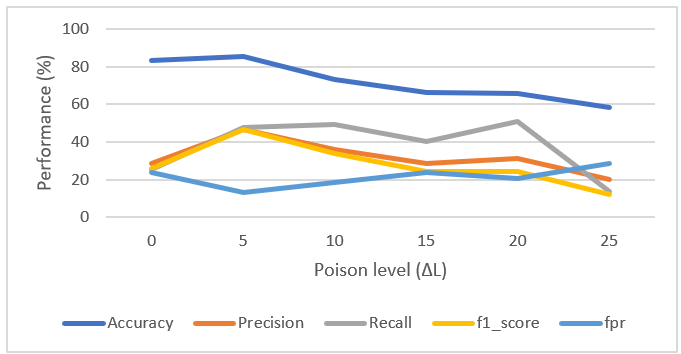}}
\caption{Performance analysis of Support Vector Machines(SVM) with consistent poisoning}
\label{fig:Performance analysis of Support Vector Machines(SVM) with consistent poisoning}
\end{figure*}

\begin{figure*}[!t]
\centering
\subfigure[\small Poisoning RF with IRIS dataset]
    {\includegraphics[width=5.8cm,height=3.1cm]{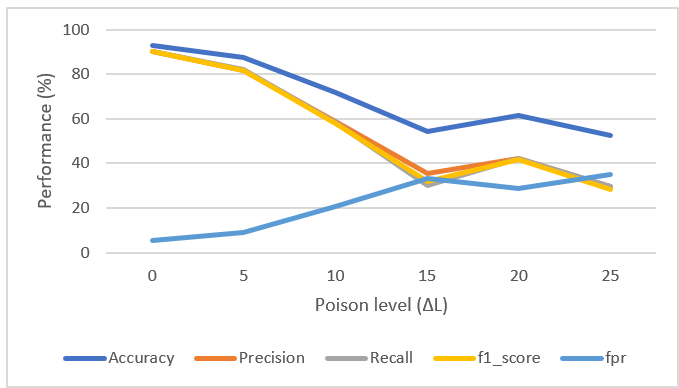}}
\subfigure[\small Poisoning RF with MNIST dataset]
    {\includegraphics[width=5.8cm,height=3.1cm]{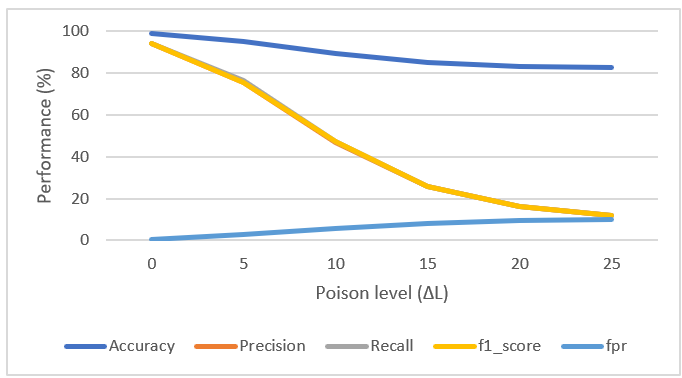}}
\subfigure[\small Poisoning RF with ISIC dataset]
    {\includegraphics[width=5.8cm,height=3.1cm]{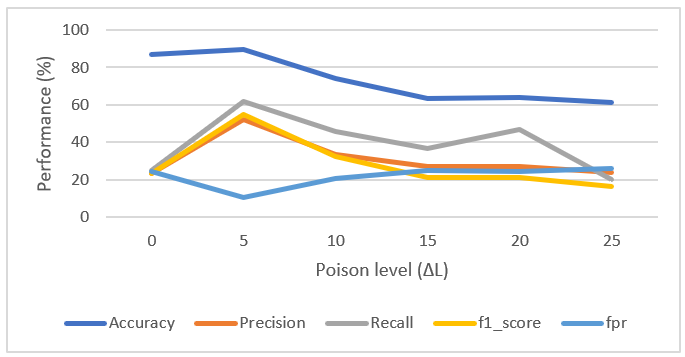}}
\caption{Performance analysis of Random Forest(RF) with consistent poisoning}
\label{fig:Performance analysis of Random Forest(RF) with consistent poisoning}
\end{figure*}

\begin{figure*}[!t]
\centering
\subfigure[\small Poisoning GNB with IRIS dataset]
    {\includegraphics[width=5.8cm,height=3.1cm]{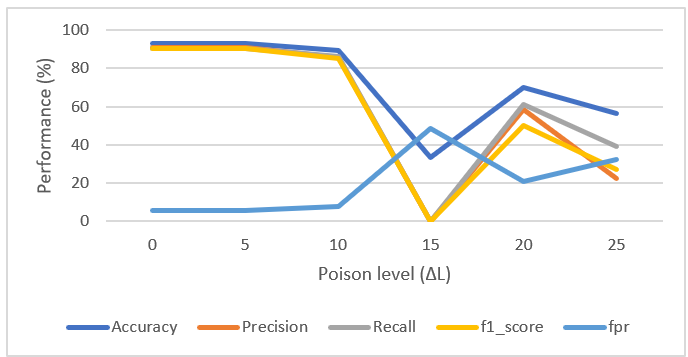}}
\subfigure[\small Poisoning GNB with MNIST dataset]
    {\includegraphics[width=5.8cm,height=3.1cm]{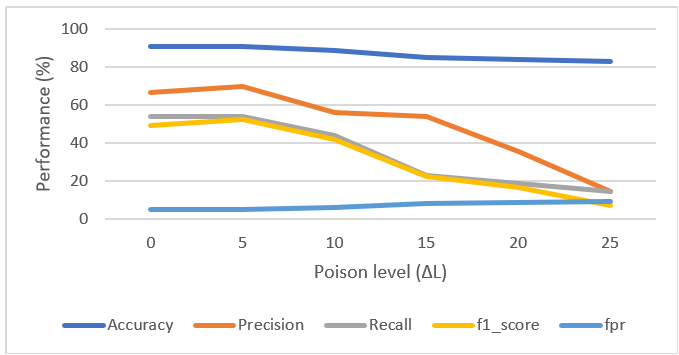}}
\subfigure[\small Poisoning GNB with ISIC dataset]
    {\includegraphics[width=5.8cm,height=3.1cm]{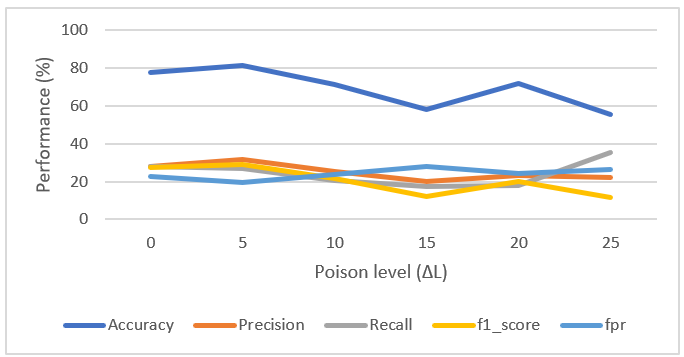}}
\caption{Performance analysis of Gaussian Naive Bayes(GNB) with consistent poisoning}
\label{fig:Performance analysis of Gaussian Naive Bayes(GNB) with consistent poisoning}
\end{figure*}

\begin{figure*}[!t]
\centering
\subfigure[\small Poisoning KNN with IRIS dataset]
    {\includegraphics[width=5.8cm,height=3.1cm]{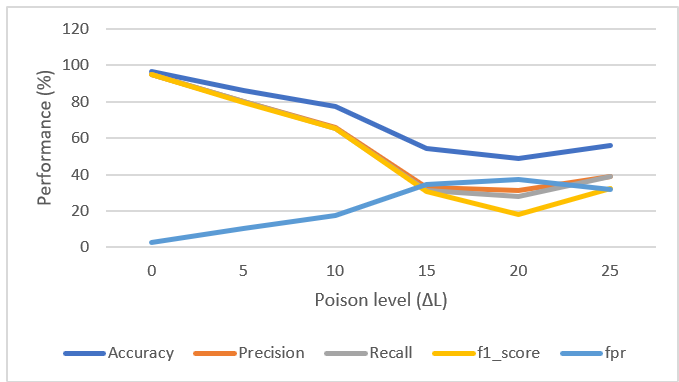}}
\subfigure[\small Poisoning KNN with MNIST dataset]
    {\includegraphics[width=5.8cm,height=3.1cm]{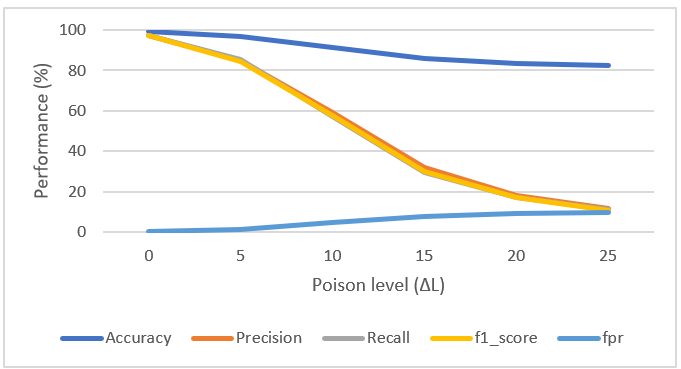}}
\subfigure[\small Poisoning KNN with ISIC dataset]
    {\includegraphics[width=5.8cm,height=3.1cm]{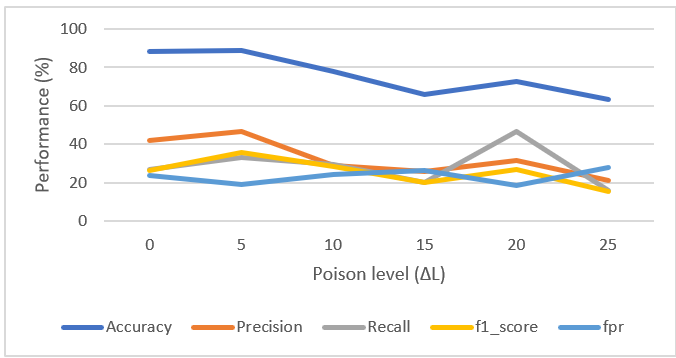}}
\caption{Performance analysis of K-Nearest Neighbours(KNN) with consistent poisoning}
\label{fig:Performance analysis of K-Nearest Neighbours(KNN) with consistent poisoning}
\end{figure*}

\begin{figure*}[!t]
\centering
\subfigure[\small Poisoning DT with IRIS dataset]
    {\includegraphics[width=5.8cm,height=3.1cm]{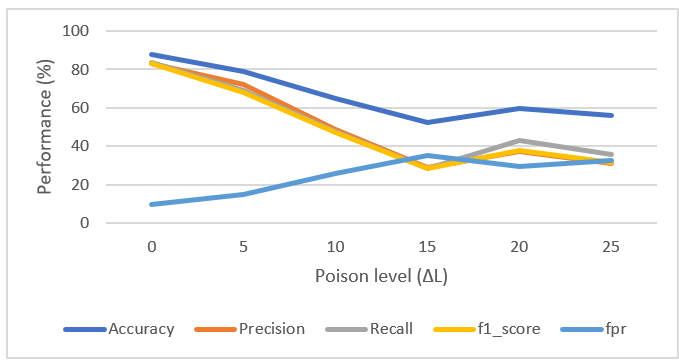}}
\subfigure[\small Poisoning DT with MNIST dataset]
    {\includegraphics[width=5.8cm,height=3.1cm]{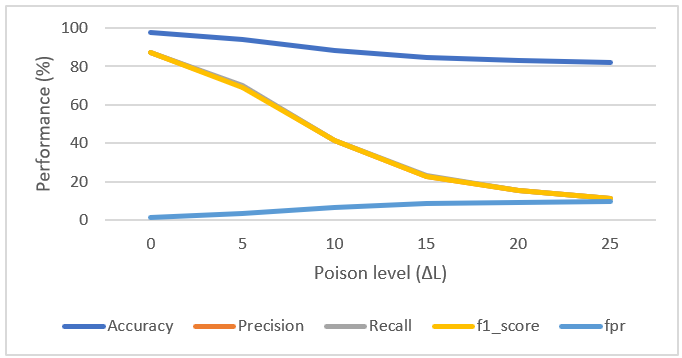}}
\subfigure[\small Poisoning DT with ISIC dataset]
    {\includegraphics[width=5.8cm,height=3.1cm]{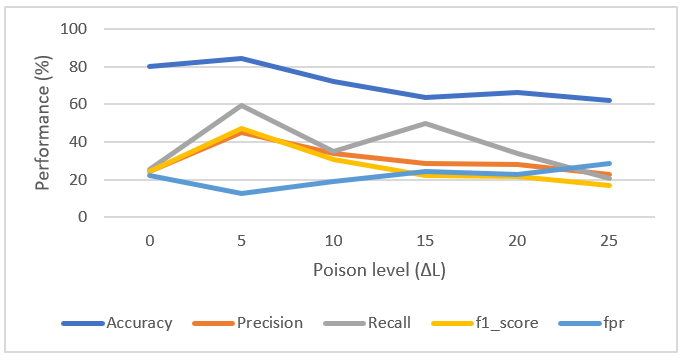}}
\caption{Performance analysis of Decision Tree(DT) with consistent poisoning}
\label{fig:Performance analysis of Decision Tree(DT) with consistent poisoning}
\end{figure*}

\begin{figure*}[!t]
\centering
\subfigure[\small Poisoning MLP with IRIS dataset]
    {\includegraphics[width=5.8cm,height=3.1cm]{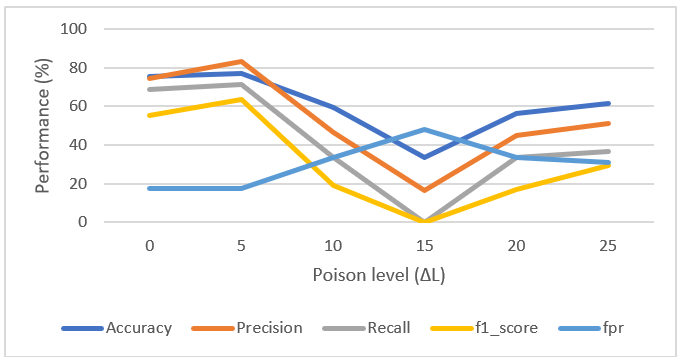}}
\subfigure[\small Poisoning MLP with MNIST dataset]
    {\includegraphics[width=5.8cm,height=3.1cm]{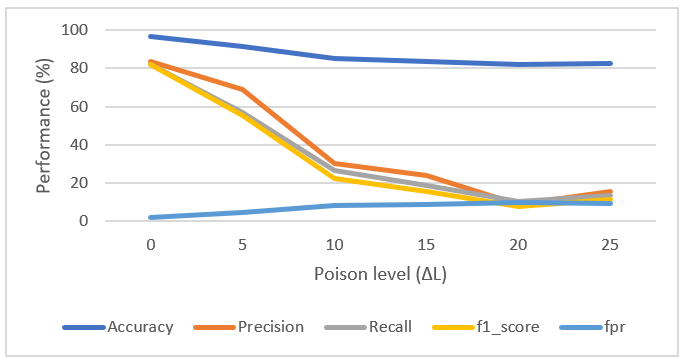}}
\subfigure[\small Poisoning MLP with ISIC dataset]
    {\includegraphics[width=5.8cm,height=3.1cm]{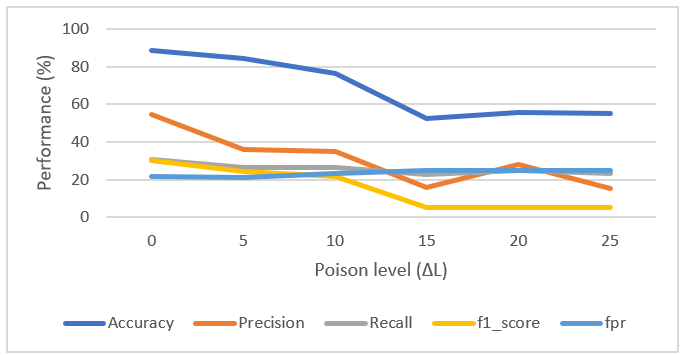}}
\caption{Performance analysis of MLP with consistent poisoning}
\label{fig:Performance analysis of Perceptron with consistent poisoning}
\end{figure*}

We have initially evaluated the impact of our OOP attack on various multiclass models using three datasets to degrade their overall performance with the attack settings given in Section \ref{outlier-oriented Poisoning (OOP) Attack Settings}. Baseline results are given in Figure \ref{fig:Performance analysis of Support Vector Machines(SVM) with consistent poisoning} to Figure \ref{fig:Performance analysis of Perceptron with consistent poisoning} where validation accuracy, precision, recall, f1-score and False Positive Rate are plotted against poisoned training with maximum poisoning level $\Delta L=25\%$. Our findings indicate that the KNN algorithm was particularly vulnerable, experiencing the most significant accuracy disruption with a maximum decrease in accuracy $(\lambda)=40.35$ at $\Delta L=25\%$ with an increase in FPR=31.6\% from FPR=2.7\%, shown in Figure \ref{fig:Performance analysis of K-Nearest Neighbours(KNN) with consistent poisoning}(a). This vulnerability stems from KNN being a non-parametric algorithm that relies on the proximity of data points to determine class features. Our attack manipulates these features spaces exploiting their outliers, and causing misclassification. From Table \ref{Analyzing k affecting KNN poisoning}, no. of nearest neighbors found to be inversely proportional to $\Delta L$, reducing attack success rate 15.79\% to 2.76\% for IRIS dataset by changing k=3 to k=15. Figure \ref{fig:Performance analysis of K-Nearest Neighbours(KNN) with consistent poisoning}(c) demonstrates high ASR when KNN is trained with ISIC dataset, decreasing its validation accuracy to 63\% with FPR=28.25\%. Where from Table \ref{Analyzing k affecting KNN poisoning}, increasing no. of nearest neighbors decreases ASR from 3.97\% to 3.31\% with $\Delta P=25\%$.
\begin{table}[h]
\small
    \centering
\captionsetup[subfloat]{width=0.5\linewidth, justification=centering}
    \caption{Analyzing k-neighbors affecting KNN accuracy with $\Delta L=(0, 10, 15, 25)\%$}
    \label{Analyzing k affecting KNN poisoning}
    \begin{tabular}{c c c c c c}
    \toprule
         & Poison Level &  k=3 & k=5 & k=10 & k=15\\
    \midrule
         \multirow{4}{*}{IRIS} & $\Delta L=0\%$ & 94.73 & 97.50 & 97.36 & 97.36\\
         & $\Delta L=10\%$ & 89.47 & 97.36 & 97.30 & 94.73\\
         & $\Delta L=15\%$ & 81.57 & 92.10 & 94.73 & 92.10\\
         & $\Delta L=20\%$ & 78.94 & 84.21 & 94.60 & 94.60\\
         \hline
         \multirow{4}{*}{MNIST} & $\Delta L=0\%$ & 98.16 & 97.55 & 96.94 & 96.55\\
         & $\Delta L=10\%$ & 92.41 & 96.52 & 96.78 & 96.50\\
         & $\Delta L=15\%$ & 89.44 & 90.90 & 94.54 & 95.95\\
         & $\Delta L=25\%$ & 85.34 & 76.14 & 83.68 & 87.52\\
         \hline
         \multirow{4}{*}{ISIC} & $\Delta L=0\%$ & 80.79 & 82.11 & 70.19 & 77.48\\
         & $\Delta L=10\%$ & 77.48 & 77.48 & 66.88 & 74.17\\
         & $\Delta L=15\%$ & 76.15 & 74.17 & 68.87 & 76.13\\
         & $\Delta L=25\%$ & 76.82 & 74.07 & 64.90 & 74.17\\
         \bottomrule 
    \end{tabular}
\end{table}

GNB is the second most affected algorithm with a decrease in validation accuracy from 92.98\% to 56.14\% and an increase in FPR from 5.68\% to 32.49\% at $0\% \leq \Delta L \leq 25\%$, for IRIS dataset, given in Fig \ref{fig:Performance analysis of Gaussian Naive Bayes(GNB) with consistent poisoning}. Interestingly, the GNB model is failing with our attack at $\Delta L=15\%$ where it's $precision\leq 0$ where a lower impact can be seen with MNIST and ISIC datasets. Further analysis reveals the change in the importance of classes, leading to misclassification, with changing class probabilities at poisoning levels, given in Table \ref{Analyzing class probabilities of GNB}. Our attack manipulates the Gaussian probability measures, making the highest probability class an anomaly and vice versa for the IRIS dataset however minor changes are visible for MNIST and ISIC datasets with no change in classes ranking at $0\% \leq \Delta L \leq 15\%$. Our analysis also highlights that GNB is the most affected algorithm when trained with a dataset with fewer classes.
\begin{table}[h]
\small
    \centering
\captionsetup[subfloat]{width=0.5\linewidth, justification=centering}
    \caption{Analyzing class probabilities of GNB with poisoned dataset}
    \label{Analyzing class probabilities of GNB}
    \begin{tabular}{p{3em} p{4em} p{4em} p{5em} p{5em}}
\toprule
Dataset & Dataset Class & Clean Dataset & $\Delta L=10\%$ & $\Delta L=15\%$\\
    \midrule
    \multirow{3}{*}{IRIS} & Class 0 & 0.33 & 0.36 & 0.38\\
    & Class 1 & 0.35 & 0.25 & 0.33\\
    & Class 2 & 0.31 & 0.37 & 0.27\\\hline
    \multirow{10}{*}{MNIST} & Class 0 & 0.09 & 0.09 & 0.09\\
    & Class 1 & 0.11 & 0.11 & 0.11\\
    & Class 2 & 0.09 & 0.09 & 0.09\\
    & Class 3 & 0.10 & 0.10 & 0.10\\
    & Class 4 & 0.09 & 0.10 & 0.09\\
    & Class 5 & 0.08 & 0.09 & 0.09\\
    & Class 6 & 0.09 & 0.09 & 0.09\\
    & Class 7 & 0.10 & 0.10 & 0.10\\
    & Class 8 & 0.09 & 0.09 & 0.09\\
    & Class 9 & 0.09 & 0.10 & 0.10\\\hline
    \multirow{4}{*}{ISIC} & Class 0 & 0.76 & 0.69 & 0.64\\
    & Class 1 & 0.05 & 0.08 & 0.10\\
    & Class 2 & 0.02 & 0.04 & 0.07\\
    & Class 3 & 0.14 & 0.17 & 0.17\\
    \bottomrule 
    \end{tabular}
\end{table}
 Whereas our OOP attack has minimally disrupted DT, resulting in $\lambda$ values of 31.6 for IRIS, 15.18 for MNIST, and 17.88 for ISIC at $\Delta L=25\%$. Table \ref{Features Importance Scores - Decision Tree} demonstrates the change in features importance scores with dataset poisoning where feature1 scores (0.90, 0.39) remain highest for IRIS and MNIST. But feature1 (0.36) with the highest importance score for ISIC becomes anomalous making anomaly feature2 (0.37), the most important feature at $\Delta L=15\%$, degrading its classification. Random Forest algorithm demonstrates relative robustness, with its FPR converge to $\approx2\%$ with an overall accuracy decrease to 61.25\% from 87\% for the ISIC dataset and FPR converge to $\approx9\%$ for the MNIST dataset with accuracy of 82.38\% at $\Delta L=25\%$ as shown in Figure \ref{fig:Performance analysis of Random Forest(RF) with consistent poisoning}. Because RF follows the ensemble approach and classifies averaging decisions from all of its trees which normalizes the poisoning effects in our case. The change in features importance scores for RF is given in Table \ref{Features Importance Scores - Random Forest} where features ranks remain the same for IRIS and MNIST but for ISIC highest ranked feature dropped to rank two at $\Delta L=15\%$ poisoning. Lastly, SVM and MLP are also not found to be very sensitive to our OOP attack. For SVM, features ranks remain intact, given in Table \ref{Features Importance Scores - SVM}, except for ISIC where feature3 (0.39) importance score reduces to (0.33) at $\Delta L=15\%$ making it an anomaly. A lower impact is visible on MLP from Fig \ref{fig:Performance analysis of Perceptron with consistent poisoning}, with our attack except at $\Delta L=15\%$ where it is failing for the IRIS dataset.

\begin{table*}[h]
\small
    \centering
    \caption{Features importance score - DT where $\Delta L=(0\%, 10\%, 15\%)$}
    \label{Features Importance Scores - Decision Tree}
    \begin{tabular}{c c c c c c c c c c}
\toprule
\multirow{3}{*}{Dataset} & \multicolumn{3}{c }{Clean Dataset} & \multicolumn{3}{ c }{Poisoned Dataset $\Delta L=10\%$} & \multicolumn{3}{ c}{Poisoned Dataset $\Delta L=15\%$}\\
\cline{2-4} \cline{5-7} \cline{8-10}
& Feature1 & Feature2 & Feature3 & Feature1 & Feature2 & Feature3 & Feature1 & Feature2 & Feature3 \\
\midrule
IRIS & \cellcolor[HTML]{C47B12}0.90 & 0.00 & 0.02 & \cellcolor[HTML]{F4AF52}0.87 & 0.008 & 0.11 & \cellcolor[HTML]{F2CB97}0.79 & 0.07 & 0.12\\
MNIST & \cellcolor[HTML]{C47B12}0.39 & 0.34 & 0.26 & \cellcolor[HTML]{F4AF52}0.39 & 0.33 & 0.27 & \cellcolor[HTML]{F4AF52}0.39 & 0.32 & 0.28 \\
ISIC & \cellcolor[HTML]{C47B12}0.36 & 0.28 & 0.35 & \cellcolor[HTML]{F2CB97}0.28 & 0.38 & 0.32 & \cellcolor[HTML]{F4AF52}0.32 & 0.37 & 0.30\\
\bottomrule 
    \end{tabular}
\end{table*}

\begin{table*}[h]
\small
    \centering
    \caption{Features importance score - SVM where $\Delta L=(0\%, 10\%, 15\%)$}
    \label{Features Importance Scores - SVM}
    \begin{tabular}{c c c c c c c c c c}
\toprule
\multirow{3}{*}{Dataset} & \multicolumn{3}{c }{Clean Dataset} & \multicolumn{3}{ c }{Poisoned Dataset $\Delta L=10\%$} & \multicolumn{3}{ c}{Poisoned Dataset $\Delta L=15\%$}\\
\cline{2-4} \cline{5-7} \cline{8-10}
& Feature1 & Feature2 & Feature3 & Feature1 & Feature2 & Feature3 & Feature1 & Feature2 & Feature3 \\
\midrule
IRIS & \cellcolor[HTML]{C47B12}0.90 & 0.02 & 0.08 & \cellcolor[HTML]{F2CB97}0.78& 0.05 & 0.15 & \cellcolor[HTML]{F4AF52}0.86 & 0.10 & 0.02\\
MNIST & 0.40 & 0.16 & \cellcolor[HTML]{C47B12}0.43 & 0.34 & 0.23 & \cellcolor[HTML]{F4AF52}0.42 & 0.36 & 0.21 & \cellcolor[HTML]{F4AF52}0.42\\
ISIC & 0.33 & 0.27 & \cellcolor[HTML]{F4AF52}0.39 & 0.30 & 0.22 & \cellcolor[HTML]{C47B12}0.47 & 0.32 & 0.33 & \cellcolor[HTML]{F2CB97}0.33\\
\bottomrule 
    \end{tabular}
\end{table*}

\begin{table}[ht]
\small
    \centering
    \captionsetup{width=\linewidth}
    \caption{\\Analyzing SVM margin score for different datasets with $\Delta L= (0, 10, 15)\%$}
    \label{Minimum margin score of SVM}
    \begin{tabular}
    {c c c c}
    \toprule
         Dataset & $\Delta L=0\%$ & $\Delta L=10\%$ & $\Delta L=15\%$\\
    \midrule
         IRIS & 0.005 & 0.01 & 0.001\\
         MNIST & 0.0000011 & 0.00000022 & 0.00000027\\
         ISIC & 0.01 & 0.003 & 0.003\\
         \bottomrule 
    \end{tabular}
\end{table}

\begin{table*}[h]
\small
    \centering
    \caption{\\Features importance score - RF where $\Delta L=(0\%, 10\%, 15\%)$}
    \label{Features Importance Scores - Random Forest}
    \begin{tabular}{c c c c c c c c c c}
\toprule
\multirow{3}{*}{Dataset} & \multicolumn{3}{c }{Clean Dataset} & \multicolumn{3}{ c }{Poisoned Dataset $\Delta L=10\%$} & \multicolumn{3}{ c}{Poisoned Dataset $\Delta L=15\%$}\\
\cline{2-4} \cline{5-7} \cline{8-10}
& Feature1 & Feature2 & Feature3 & Feature1 & Feature2 & Feature3 & Feature1 & Feature2 & Feature3 \\
\midrule
IRIS & \cellcolor[HTML]{C47B12}0.66 & 0.15 & 0.17 & \cellcolor[HTML]{F4AF52}0.58 & 0.19 & 0.22 & \cellcolor[HTML]{F2CB97}0.52 & 0.22 & 0.25\\
MNIST & \cellcolor[HTML]{C47B12}0.39 & 0.34 & 0.26 & \cellcolor[HTML]{C47B12}0.39 & 0.33 & 0.27 & \cellcolor[HTML]{C47B12}0.39 & 0.32 & 0.27\\
ISIC & 0.31 & \cellcolor[HTML]{F4AF52}0.35 & 0.34 & 0.31 & \cellcolor[HTML]{C47B12}0.36 & 0.32 & \cellcolor[HTML]{F2CB97}0.34 & 0.33 & 0.32\\
\bottomrule 
    \end{tabular}
\end{table*}

\subsubsection{Effects of Increasing Poisoning Rate}
We extended our analysis to study the effects of consistently increasing poisoning rates on multiclass models with our OOP attack. The aggregated results, given in Figure \ref{fig:Performance analysis of Support Vector Machines(SVM) with consistent poisoning} to Figure \ref{fig:Performance analysis of Perceptron with consistent poisoning}, shows over-fitting? \textbf{No}. Our results demonstrated that the classification accuracy of multiclass classifiers has maximum disruption when the training dataset is poisoned with our OOP attack at $\Delta L=10\%$ irrespective of datasets. An inverse relationship was observed between the number of classes in the dataset and the rate of performance degradation. For the MNIST dataset, from Fig \ref{fig:Performance analysis of Support Vector Machines(SVM) with consistent poisoning}(b) to Fig \ref{fig:Performance analysis of Perceptron with consistent poisoning}(b), with ten dataset classes has a steady decrease in performance. Whereas, classifiers trained with the IRIS dataset, with three dataset classes, have high fluctuation in performance followed by ISIC with four classes. Least percentage of data poisoning is more effective on parametric models. 10\% poisoning has a steady and practical impact on parametric models whereas 15\% poisoning leads to impractical effects. From Fig \ref{fig:Performance analysis of Gaussian Naive Bayes(GNB) with consistent poisoning}(a) and Fig \ref{fig:Performance analysis of Perceptron with consistent poisoning}(a) parametric models, with minimum no. of classes, are failing at 15\% poisoning. But, $\Delta L=15\%$ is very effective for non-parametric models. Conclusively, $10\% \leq \Delta L \leq 15\%$ are the optimal poisoning rates for multiclass models where $\Delta L>15\%$ shows an impractical success.

\subsubsection{Model Sensitivity to Poison and Effects of Data Distribution}
\begin{figure*}[!t]
\centering
\subfigure[\small Data Distribution of IRIS Dataset with OOP Attack]
    {\includegraphics[width=7.8cm,height=5.2cm]{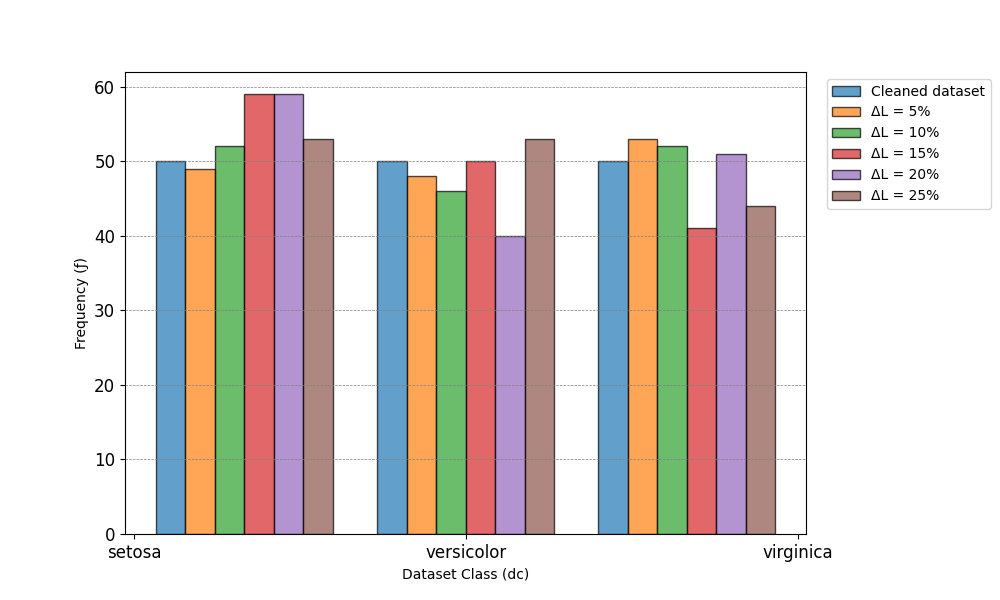}}
\subfigure[\small Data Distribution of MNIST Dataset with OOP Attack]
    {\includegraphics[width=7.8cm,height=5.2cm]{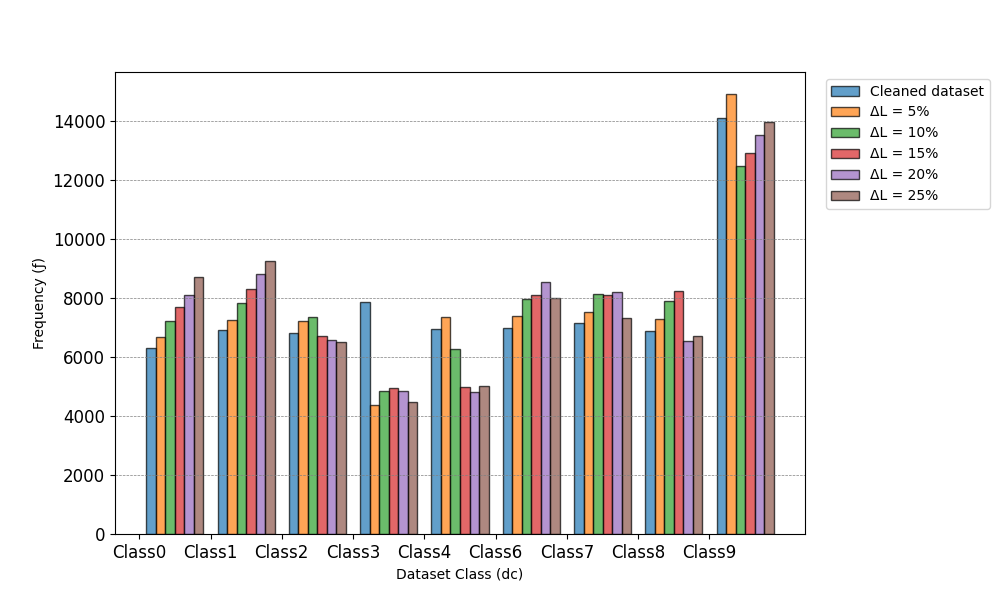}}
\subfigure[\small Data Distribution of ISIC Dataset with Distance-based Attack]
    {\includegraphics[width=7.8cm,height=5.2cm]{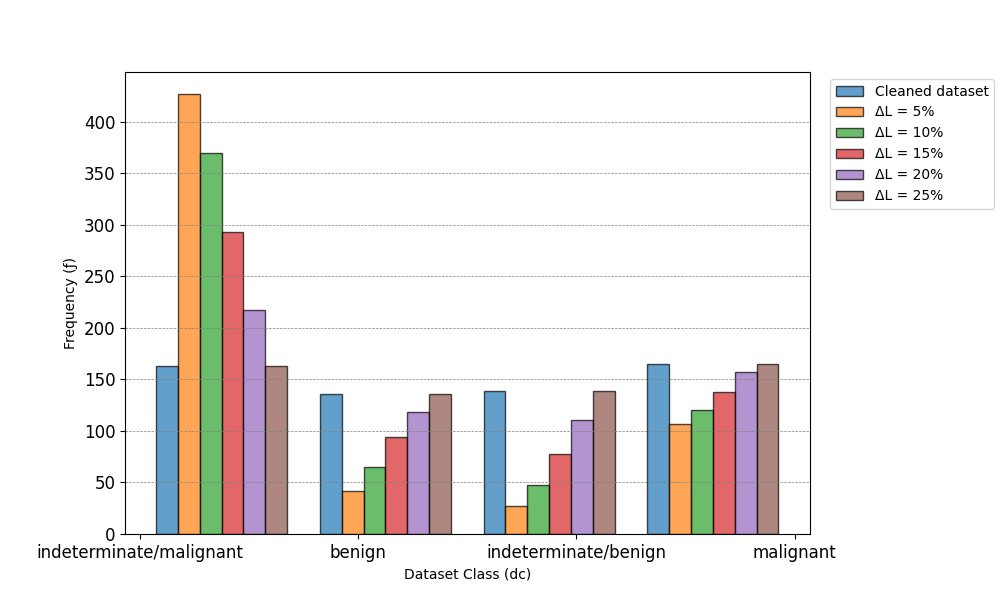}}
\caption{Data Distribution with Distance-based Attack}
\label{Data Distribution with Distance-based Attack}
\end{figure*}

We investigated model sensitivity by analyzing the relationship between model variance and ASR. Table \ref{Variance in Algorithm at Different Poisoning Levels} illustrates the variance in machine learning models in response to our OOP attack. This attack significantly increased the sensitivity of all tested models, with GNB exhibiting the highest sensitivity. Its variance leads to 0.8 at $\Delta L=10\%$, for the IRIS dataset, almost equivalent to DT where it fails. Similarly, 0.10 variance increases for KNN at $\Delta L=15\%$, highlighting its high sensitivity and effectiveness of our OOP attack. And, RF and DT are proved to be less sensitive to our outlier-oriented attack. Interestingly, on average models trained with MNIST and ISIC are also less affected by our poisoning attack compared to models trained with IRIS dataset with high impact.\\ 
\begin{table}[h]
\small
    \centering
\captionsetup[subfloat]{width=0.5\linewidth, justification=centering}
    \caption{\\Model Variance at Different Poisoning Levels}
    \label{Variance in Algorithm at Different Poisoning Levels}
    \begin{tabular}{c c c c c}
\toprule
Dataset & Algorithm & Clean Dataset & $\Delta L=10\%$ & $\Delta L=15\%$\\
    \midrule
    \multirow{6}{*}{IRIS} & SVM & 0.33 & 0.36 & 0.57\\
    & RF & 0.62 & 0.60 & 0.63\\
    & GNB & 0.65 & 0.73 & 0.68\\
    & KNN & 0.81 & 0.82 & 0.91\\
    & DT & 0.59 & 0.68 & 0.78\\
    & MLP & 0.65 & 0.69 & 1.45\\\hline
    \multirow{6}{*}{MNIST} & SVM & 8.33 & 8.06 & 7.97\\
    & RF & 8.24 & 7.69 & 7.71\\
    & GNB & 11.25 & 12.68 & 12.74\\
    & KNN & 8.36 & 8.37 & 8.38\\
    & DT & 8.33 & 7.81 & 8.02\\
    & MLP & 8.31 & 8.38 & 8.31\\\hline
    \multirow{6}{*}{ISIC} & SVM & 1.33 & 0.97 & 1.36\\
    & RF & 1.11 & 1.17 & 1.32\\
    & GNB & 1.27 & 1.66 & 1.19\\
    & KNN & 0.31 & 0.37 & 0.27\\
    & DT & 0.31 & 0.37 & 0.27\\
    & MLP & 1.52 & 1.48 & 1.59\\
    \bottomrule 
    \end{tabular}
\end{table}
Further analysis was conducted on dataset distribution to ascertain its impact on data poisoning and performance degradation in models. Fig \ref{Data Distribution with Distance-based Attack} shows the change in data distribution with our OOP attack at $0\% \leq \Delta L \leq 25\%$. Our findings suggest that balanced datasets with a greater number of classes tend to mitigate the effects of poisoning on model performance, particularly in terms of model accuracy. In contrast, imbalanced and noisy datasets work as catalysts and boost the poisoning effects of our attack, leading to impractically high decrease in performance such as for the ISIC dataset in our case,  as shown in Fig \ref{Data Distribution with Distance-based Attack}(c). From our analysis, we have identified relations of various classification characteristics and subsequent rates of data poisoning in Table \ref{Analyzing one-to-one parametric relation between poison and machine learning algorithms}.

\begin{table}[h]
\small
    \centering
    \captionsetup[subfloat]{width=0.5\linewidth, justification=centering}
    \caption{\\Analyzing one-to-one relation between poison and various parameters of machine learning algorithms}
    \label{Analyzing one-to-one parametric relation between poison and machine learning algorithms}
    \begin{tabular}{c c c}
    \toprule
         Algorithm & Algorithmic Parameters & Relation to $\Delta P$\\
    \midrule
         \multirow{3}{*}{SVM} & Margin score & Minimal impact\\
         & Decision boundary & Minimal impact\\
         & Features importance score & Minimal impact\\\hline
         \multirow{2}{*}{DT} & Features importance score & Minimal impact\\
         & Asymmetric features space & High impact\\\hline
         \multirow{2}{*}{KNN} & Decision boundary & High impact\\
         & k-neighbors & Inverse impact\\\hline
         \multirow{2}{*}{GNB} & Decision boundary & High impact\\
         & Class probabilities & High impact\\\hline
         RF & No. of trees & Inverse impact\\
         & Features importance score & Minimal impact\\\hline
         MLP & Weights & High impact\\
         \bottomrule 
    \end{tabular}
\end{table}
\section{Discussion and Limitations}
\begin{itemize}
\item \textbf{Our outlier-oriented poisoning (OOP) attack method} We formalize a novel grey-box attack to attempt poisoning multiclass models, describing their efficacy and analyzing the factors affecting their classification behavior. Although, several adversarial poisoning techniques are proposed in the literature, but limited experimentation is provided on multiclass classifiers. Existing research papers \cite{steinhardt2017certified}, \cite{chen2023tutorial}, \cite{hayase2021spectre} and \cite{weerasinghe2021defending} proposed solutions focusing discrete dataset features and detecting outliers to lessen poisoning effects. We have taken the outliers into the feature space to effectively poison the model. Following this, we have highlighted certain factors affecting individual algorithms and also determined effective levels of poisoning for parametric and non-parametric multiclass models. Our results showed that a 10\% poisoning rate is optimal for the parametric and 15\% for the non-parametric models. At these optimal poisoning, we have analyzed a lower level of model sensitivity which does not allow the model to over-fit, highlighting the efficacy of our attack. 
\item \textbf{Factors affecting the behavior of poisoned multiclass models} Implementing OOP attack, we have conducted a deep behavioral analysis of multiclass machine learning, identifying factors affecting the confidence of models. From our results, GNB and KNN are found to be highly affected by our poisoning attack whereas DT and RF are less affected models. Manipulating the outliers class label, class probabilities of GNB, and proximity distance calculation of KNN are highly disrupted. Conversely, RF and DT are attack-agnostic algorithms because of their resilience against outliers.
\item \textbf{Impact of dataset structure on multiclass model poisoning} Our results highlight that the dataset size and its no. of classes are inversely proportional to poisoning effects. Whereas, an accelerating impact of an imbalanced dataset on model poisoning. Imbalanced classes in multiclass datasets help penetrate poison in the model effectively, to an extent. Also, a fundamental relation between dataset noise and data poisoning is found where dataset noise works as a catalyst towards poisoning leading to more adverse but impractical performance degradation. 
\item \textbf{Limitations} Our research is limited to the analysis of classification algorithms which can be extended to the regression algorithms. With this limitation, we have analyzed the factors affecting classification behaviors and their confidence in our poisoning attack. Also, comparing our attack with our existing attacks from the literature helps demonstrate the efficacy of our attack which is also out of the scope of this study. 
\end{itemize}
\section{Conclusion and Future Work} This paper analyzes the behavior of multiclass machine learning models, identifying individual characteristics of the algorithms against data outlier-oriented poisoning. We formulated outlier-oriented poisoning to compromise algorithm classification in multiclass settings. Our research analyzed the sensitivity of individual algorithms against the OOP attack, identifying their key characteristics. For example, change in decision boundaries is highly disrupted in KNN and GNB, but minimum effects are visible in SVM. Features importance scores of SVM and RF have limited impact, and no. of trees in RF and no. of k-neighbors in KNN have shown an inverse impact with increasing poison rates. Further analysis has resulted in identifying the most effective poisoning rates, i.e. 10\% poisoning for the parametric and 15\% for the non-parametric algorithms and impractical impact of poisoning $>15\%$ with a high and fluctuating decrease in performance without overfitting. Our results showed that KNN and GNB are the most affected algorithms, whereas RF and DT are resilient against OOP attacks. Our analysis also highlights that the noisy datasets with non-uniform features aggravate the poisoning effects. In contrast, dense datasets with higher no. of classes normalize the poisoning effects, particularly the accuracy of the models.\\
For future work following our behavioral analysis, some potential directions to mitigate these poisoning attacks are: 
\begin{itemize}
\item Improve adversarial training of the models by training against 10\%-15\% of the poisoned dataset as these are the effective poisoning levels.
\item Model hardening can also be better implemented following the effects on individual model parameters such as k-neighbors in KNN, support vectors in SVM, etc with ensemble learning settings rather than following general performance metrics of the models such as accuracy, precision, recall, and FPR.
\item To remediate the outliers or anomalies focused attacks, we can develop pre-training models to identify outliers and cleanse our dataset.
\end{itemize}

\textbf{CRediT authorship contribution statement}\\\\
\textbf{Anum Paracha} - Problem Statement, Conceptualization, Investigation, Formal Analysis, Writing - Original draft. \textbf{Junaid Arshad} - Conceptualization, Writing - Original draft. \textbf{Mohamed Ben Farah} - Conceptualization, Writing - Review and Update. \textbf{Khalid Ismail} - Conceptualization, Writing - Review and Update.\\\\
\textbf{Declaration of competing interest}\\\\
The authors declare that they have no known competing financial interests or personal relationships that could have appeared
to influence the work reported in this paper.\\\\
\textbf{Data availability}\\\\
Data will be made available upon request.\\\\
\textbf{Acknowledgement}\\\\
No acknowledgments to declare.


\begin{thebibliography}{00}
\bibitem{zhou2024object}
\href{https://www.sciencedirect.com/science/article/pii/S2214212624000139}{Zhou, C., Wang, Y. \& Zhu, G. Object-attentional untargeted adversarial attack. {\em Journal Of Information Security And Applications}. \textbf{81} pp. 103710 (2024)}

\bibitem{schneider2023dual}
\href{https://www.sciencedirect.com/science/article/pii/S2214212623000868}{Schneider, J. \& Apruzzese, G. Dual adversarial attacks: Fooling humans and classifiers. {\em Journal Of Information Security And Applications}. \textbf{75} pp. 103502 (2023)}

\bibitem{mccarthy2023defending}
\href{https://www.sciencedirect.com/science/article/pii/S2214212622002423}{McCarthy, A., Ghadafi, E., Andriotis, P. \& Legg, P. Defending against adversarial machine learning attacks using hierarchical learning: A case study on network traffic attack classification. {\em Journal Of Information Security And Applications}. \textbf{72} pp. 103398 (2023)}

\bibitem{aghakhani2021bullseye}
\href{https://doi.org/10.1109/EuroSP51992.2021.00021}{Hojjat Aghakhani et al. “Bullseye polytope: A scalable
clean-label poisoning attack with improved
transferability”. In: \emph{IEEE European symposium
on security and privacy (EuroS\&P)}. IEEE. 2021,
pp. 159–178.}

\bibitem{alarab2023uncertainty}
\href{https://doi.org/10.1007/s11042-022-13269-1}{Ismail Alarab and Simant Prakoonwit. “Uncertainty
estimation based adversarial attack in multiclass
classification”. In: \emph{Multimedia Tools and
Applications} 82.1 (2023), pp. 1519–1536.}

\bibitem{ali2023adversarial}
\href{https://www.sciencedirect.com/science/article/pii/S0167739X22003375}{Ali, Y. Adversarial attacks on deep learning networks in image classification based on Smell Bees Optimization Algorithm. {\em Future Generation Computer Systems}. \textbf{140} pp. 185-195 (2023)}

\bibitem{li2022query}
\href{https://www.sciencedirect.com/science/article/pii/S0167739X2200084X}{Li, S., Huang, G., Xu, X. \& Lu, H. Query-based black-box attack against medical image segmentation model. {\em Future Generation Computer Systems}. \textbf{133} pp. 331-337 (2022)}

\bibitem{aryal2022analysis}
\href{https://doi.org/10.1109/BigData55660.2022.10020528}{Kshitiz Aryal, Maanak Gupta, and
Mahmoud Abdelsalam. “Analysis of label-flip poisoning
attack on machine learning based malware detector”. In:
\emph{IEEE International Conference on Big Data (Big
Data)}. IEEE. 2022, pp. 4236–4245.}

\bibitem{10.5555/3042573.3042761}
\href{https://doi.org/10.5555/3042573.3042761}{Battista Biggio, Blaine Nelson, and Pavel Laskov. “Poisoning attacks against support vector machines”. In:
\emph{Proceedings of the 29th International Coference on International Conference on Machine Learning}. ICML’12. Edinburgh, Scotland: Omnipress, 2012, pp. 1467–1474. \textsc{isbn}: 9781450312851.}

\bibitem{bostani2024evadedroid}
\href{https://doi.org/10.1016/j.cose.2023.103676}{Hamid Bostani and Veelasha Moonsamy. “Evadedroid: A practical evasion attack on machine learning for black-box android malware detection”. In: \emph{Computers \& Security} 139 (2024), p. 103676.}

\bibitem{carlini2021poisoning}
\href{https://arxiv.org/abs/2106.09667}{Nicholas Carlini and Andreas Terzis. “Poisoning and backdooring contrastive learning”. In: \emph{arXiv preprint arXiv:2106.09667} (2021).}

\bibitem{chen2023tutorial}
\href{https://doi.org/10.1145/3574159}{Huili Chen and Farinaz Koushanfar. “Tutorial: toward
robust deep learning against poisoning attacks”. In: \emph{ACM Transactions on Embedded Computing Systems} 22.3 (2023), pp. 1–15.}

\bibitem{chen2021deeppoison}
\href{https://doi.org/10.1109/TCSII.2021.3060896}{Jinyin Chen et al. “Deeppoison: Feature transfer based stealthy poisoning attack for dnns”. In: \emph{IEEE Transactions on Circuits and Systems II: Express Briefs} 68.7 (2021), pp. 2618–2622.}

\bibitem{chillara2024deceiving}
\href{https://doi.org/10.1007/s10207-024-00834-y}{Anil Kumar Chillara et al. “Deceiving supervised machine learning models via adversarial data poisoning attacks: a case study with USB keyboards”. In: \emph{International Journal of Information Security} (2024), pp. 1–19.}

\bibitem{di2022hidden}
\href{https://doi.org/10.5555/3666122.3668045}{Jimmy Z Di et al. “Hidden poison: Machine unlearning enables camouflaged poisoning attacks”. In: \emph{NeurIPS ML Safety Workshop}. 2022.}

\bibitem{han2023credential}
\href{https://doi.org/10.1109/TIFS.2023.3326985}{Ruidong Han et al. “A Credential Usage Study: Flow-Aware Leakage Detection in Open-Source Projects”. In: \emph{IEEE Transactions on Information Forensics and Security} (2023).}

\bibitem{hayase2021spectre}
\href{https://proceedings.mlr.press/v139/hayase21a.html}{Jonathan Hayase et al. “Spectre: Defending against backdoor attacks using robust statistics”. In: \emph{International Conference on Machine Learning}. PMLR. 2021, pp. 4129–4139.}

\bibitem{hossain2024advancing}
\href{https://doi.org/10.48550/arXiv.2403.08208}{Khondoker Murad Hossain and Tim Oates. “Advancing Security in AI Systems: A Novel Approach to Detecting Backdoors in Deep Neural Networks”. In: \emph{arXiv preprint arXiv:2403.08208} (2024).}

\bibitem{huang2020metapoison}
\href{https://proceedings.neurips.cc/paper/2020/hash/8ce6fc704072e351679ac97d4a985574-Abstract.html}{W Ronny Huang et al. “Metapoison: Practical general-purpose clean-label data poisoning”. In: \emph{Advances in Neural Information Processing Systems} 33 (2020), pp. 12080–12091.}

\bibitem{huang2022embedding}
\href{https://doi.org/10.1007/s10994-021-06068-6}{Wei Huang, Xingyu Zhao, and Xiaowei Huang.
“Embedding and extraction of knowledge in tree ensemble classifiers”. In: \emph{Machine Learning} 111.5 (2022), pp. 1925–1958.}

\bibitem{10.1145/3460120.3485368}
\href{https://doi.org/10.1145/3460120.3485368}{Matthew Jagielski et al. “Subpopulation Data Poisoning Attacks”. In: \emph{Proceedings of the 2021 ACM SIGSAC Conference on Computer and Communications Security}. CCS ’21. Virtual Event, Republic of Korea: Association for Computing Machinery, 2021, pp. 3104–3122. \textsc{isbn}: 9781450384544.}

\bibitem{jha2024label}
\href{https://proceedings.neurips.cc/paper_files/paper/2023/hash/e0c9b65fb3e41aaa86576df3ec33ad2e-Abstract-Conference.html}{Rishi Jha, Jonathan Hayase, and Sewoong Oh. “Label
poisoning is all you need”. In: \emph{Advances in Neural Information Processing Systems} 36 (2024).}

\bibitem{jonnalagadda2024modelling}
\href{https://www.worldscientific.com/doi/abs/10.1142/S0219649224500229}{Annapurna Jonnalagadda et al. “Modelling Data Poisoning Attacks Against Convolutional Neural Networks”. In: \emph{Journal of Information “\& Knowledge Management} (2024), p. 2450022.}

\bibitem{koh2022stronger}
\href{https://doi.org/10.1007/s10994-021-06119-y}{Pang Wei Koh, Jacob Steinhardt, and Percy Liang. “Stronger data poisoning attacks break data sanitization defenses”. In: \emph{Machine Learning} (2022), pp. 1–47.}

\bibitem{laishram2016curie}
\href{https://arxiv.org/abs/1606.01584}{Ricky Laishram and Vir Virander Phoha. “Curie: A method for protecting SVM classifier from poisoning attack”. In: \emph{arXiv preprint arXiv:1606.01584} (2016).}

\bibitem{liu2022adversarial}
\href{https://link.springer.com/chapter/10.1007/978-3-031-20065-6_14}{Ganlin Liu, Xiaowei Huang, and Xinping Yi. “Adversarial Label Poisoning Attack on Graph Neural Networks via Label Propagation”. In: \emph{European Conference on Computer Vision}. Springer. 2022, pp. 227–243.}

\bibitem{liu2023gradient}
\href{https://ieeexplore.ieee.org/abstract/document/10285881}{Gaoyang Liu et al. “Gradient-Leaks: Enabling Black-Box Membership Inference Attacks Against Machine Learning Models”. In: \emph{IEEE Transactions on Information Forensics and Security} (2023).}

\bibitem{liu2023image}
\href{https://proceedings.mlr.press/v202/liu23bb.html}{Zhuoran Liu, Zhengyu Zhao, and Martha Larson. “Image shortcut squeezing: Countering perturbative availability poisons with compression”. In: \emph{International conference on machine learning}. PMLR. 2023, pp. 22473–22487.}

\bibitem{lorenz2023certifiers}
\href{https://dl.acm.org/doi/abs/10.1145/3605764.3623917}{Tobias Lorenz, Marta Kwiatkowska, and Mario Fritz. “Certifiers Make Neural Networks Vulnerable to Availability Attacks”. In: \emph{Proceedings of the 16th ACM Workshop on Artificial Intelligence and Security}. 2023, pp. 67–78.}

\bibitem{9448491}
\href{https://ieeexplore.ieee.org/abstract/document/9448491}{Ke Ma et al. “Poisoning Attack Against Estimating From Pairwise Comparisons”. In: \emph{IEEE Transactions on Pattern Analysis and Machine Intelligence} 44.10 (2022), pp. 6393–6408. \textsc{doi}: 10.1109/TPAMI.2021.3087514.}

\bibitem{manna2024trimpa}
\href{https://ieeexplore.ieee.org/abstract/document/10413593}{Debasmita Manna and Somanath Tripathy. “TriMPA: Triggerless Targeted Model Poisoning Attack in DNN”. In: \emph{IEEE Transactions on Computational Social Systems} (2024).}

\bibitem{mayerhofer2022poisoning}
\href{https://dl.acm.org/doi/abs/10.1145/3508398.3519363}{Robin Mayerhofer and Rudolf Mayer. “Poisoning attacks against feature-based image classification”. In: \emph{Proceedings of the Twelfth ACM Conference on Data and Application Security and Privacy}. 2022, pp. 358–360.}

\bibitem{melacci2021domain}
\href{https://ieeexplore.ieee.org/abstract/document/9661418}{Stefano Melacci et al. “Domain knowledge alleviates adversarial attacks in multi-label classifiers”. In: \emph{IEEE Transactions on Pattern Analysis and Machine Intelligence} 44.12 (2021), pp. 9944–9959.}

\bibitem{moradi2023recent}
\href{https://link.springer.com/article/10.1007/s12551-022-01040-7}
{Hamed Moradi et al. “Recent developments in
modeling, imaging, and monitoring of cardiovascular
diseases using machine learning”. In:
\emph{Biophysical Reviews} 15.1 (2023), pp. 19–33.}

\bibitem{munoz2017towards}
\href{https://dl.acm.org/doi/abs/10.1145/3128572.3140451}
{Luis Mu noz-Gonz alez et al. “Towards poisoning of deep learning algorithms with back-gradient optimization”.
In: \emph{Proceedings of the 10th ACM workshop on
artificial intelligence and security}. 2017, pp. 27–38.}

\bibitem{nguyen2024label}
\href{https://proceedings.neurips.cc/paper_files/paper/2023/hash/d9827e811c5a205c1313fb950c072c7d-Abstract-Conference.html}
{Bao-Ngoc Nguyen et al. “Label-Only Model Inversion
Attacks via Knowledge Transfer”. In: \emph{Advances
in Neural Information Processing Systems} 36 (2024).}

\bibitem{pantelakis2023adversarial}
\href{https://doi.org/10.1145/3600160.3605086}
{Vasileios Pantelakis et al. “Adversarial Machine
Learning Attacks on Multiclass Classification of IoT
Network Traffic”. In: \emph{Proceedings of the 18th
International Conference on Availability, Reliability and
Security}. 2023, pp. 1–8.}

\bibitem{russo2021poisoning}
\href{https://ieeexplore.ieee.org/abstract/document/9482992}
{Alessio Russo and Alexandre Proutiere. “Poisoning
attacks against data-driven control methods”. In:
\emph{American Control Conference (ACC)}. IEEE.
2021, pp. 3234–3241.}

\bibitem{saha2020hidden}
\href{https://doi.org/10.1609/aaai.v34i07.6871}
{Aniruddha Saha, Akshayvarun Subramanya, and
Hamed Pirsiavash. “Hidden trigger backdoor attacks”.
In: \emph{Proceedings of the AAAI conference on
artificial intelligence}. Vol. 34. 07. 2020,
pp. 11957–11965.}

\bibitem{10.1145/3373376.3378462}
\href{https://dl.acm.org/doi/abs/10.1145/3373376.3378462}
{Jose Rodrigo Sanchez Vicarte et al. “Game of Threads:
Enabling Asynchronous Poisoning Attacks”. In:
\emph{Proceedings of the Twenty-Fifth International
Conference on Architectural Support for Programming
Languages and Operating Systems}. ASPLOS ’20.
Lausanne, Switzerland: Association for Computing
Machinery, 2020, pp. 35–52. \textsc{isbn}:
9781450371025. \textsc{doi}: 10.1145/3373376.3378462.}

\bibitem{shahid2022label}
\href{https://ieeexplore.ieee.org/abstract/document/10022015}
{Abdur R Shahid et al. “Label flipping data poisoning
attack against wearable human activity recognition
system”. In: \emph{IEEE Symposium Series on
Computational Intelligence (SSCI)}. IEEE. 2022,
pp. 908–914.}

\bibitem{steinhardt2017certified}
\href{https://proceedings.neurips.cc/paper/2017/hash/9d7311ba459f9e45ed746755a32dcd11-Abstract.html}
{Jacob Steinhardt, Pang Wei W Koh, and Percy S Liang.
“Certified defenses for data poisoning attacks”. In:
\emph{Advances in neural information processing
systems} 30 (2017).}

\bibitem{suya2021model}
\href{https://proceedings.mlr.press/v139/suya21a.html}
{Fnu Suya et al. “Model-targeted poisoning attacks with
provable convergence”. In: \emph{International
Conference on Machine Learning}. PMLR. 2021,
pp. 10000–10010.}

\bibitem{wang2024efficient}
\href{https://arxiv.org/abs/2402.04010}
{Yihan Wang, Yifan Zhu, and Xiao-Shan Gao. “Efficient
Availability Attacks against Supervised and Contrastive
Learning Simultaneously”. In: \emph{arXiv preprint
arXiv:2402.04010} (2024).}

\bibitem{weerasinghe2021defending}
\href{https://ieeexplore.ieee.org/abstract/document/9352750}
{Sandamal Weerasinghe et al. “Defending support vector
machines against data poisoning attacks”. In:
\emph{IEEE Transactions on Information Forensics and
Security} 16 (2021), pp. 2566–2578.}

\bibitem{wongrassamee2017can}
\href{https://www.imperial.ac.uk/media/imperial-college/faculty-of-engineering/computing/public/1617-ug-projects/Vasin-Wongrassemee---Can-you-Poison-a-Machine-Learning-Algorithm.pdf}
{Vasin Wongrassamee and Luis Mu noz-Gonz alez. “Can you Poison a Machine Learning Algorithm?” In: (2017).}

\bibitem{yang2024stealthy}
\href{https://ieeexplore.ieee.org/abstract/document/10431665}
{Zhou Yang et al. “Stealthy backdoor attack for code
models”. In: \emph{IEEE Transactions on Software
Engineering} (2024).}

\bibitem{yu2024chronic}
\href{https://ojs.aaai.org/index.php/AAAI/article/view/29591}
{Fangchao Yu et al. “Chronic Poisoning: Backdoor
Attack against Split Learning”. In: \emph{Proceedings
of the AAAI Conference on Artificial Intelligence}.
Vol. 38. 15. 2024, pp. 16531–16538.}

\bibitem{zeng2024convergence}
\href{https://ieeexplore.ieee.org/abstract/document/10422883}
{Tengchan Zeng et al. “Convergence of communications,
control, and machine learning for secure and
autonomous vehicle navigation”. In: \emph{IEEE
Wireless Communications} (2024).}

\bibitem{zhang2023clean}
\href{https://doi.org/10.1016/j.ins.2023.03.124}
{Chen Zhang, Zhuo Tang, and Kenli Li. “Clean-label
poisoning attack with perturbation causing dominant
features”. In: \emph{Information Sciences} 644 (2023),
p. 118899.}

\bibitem{zhang2020online}
\href{https://proceedings.mlr.press/v120/zhang20b.html}
{Xuezhou Zhang, Xiaojin Zhu, and Laurent Lessard.
“Online data poisoning attacks”. In: \emph{Learning
for Dynamics and Control}. PMLR. 2020, pp. 201–210.}

\bibitem{zhao2022clpa}
\href{https://ojs.aaai.org/index.php/AAAI/article/view/20902}
{Bingyin Zhao and Yingjie Lao. “CLPA: Clean-label
poisoning availability attacks using generative
adversarial nets”. In: \emph{Proceedings of the AAAI
Conference on Artificial Intelligence}. Vol. 36. 8. 2022,
pp. 9162–9170.}

\bibitem{zhao2022towards}
\href{https://openaccess.thecvf.com/content/WACV2022/html/Zhao_Towards_Class-Oriented_Poisoning_Attacks_Against_Neural_Networks_WACV_2022_paper.html}
{Bingyin Zhao and Yingjie Lao. “Towards class-oriented
poisoning attacks against neural networks”. In:
\emph{Proceedings of the IEEE/CVF Winter
Conference on Applications of Computer Vision}. 2022,
pp. 3741–3750.}

\bibitem{zheng2023trojfair}
\href{https://arxiv.org/abs/2312.10508}
{Mengxin Zheng et al. “TrojFair: Trojan Fairness
Attacks”. In: \emph{arXiv preprint arXiv:2312.10508}
(2023).}

\bibitem{zhong2020backdoor}
\href{https://dl.acm.org/doi/abs/10.1145/3374664.3375751}
{Haoti Zhong et al. “Backdoor embedding in
convolutional neural network models via invisible
perturbation”. In: \emph{Proceedings of the Tenth
ACM Conference on Data and Application Security and
Privacy}. 2020, pp. 97–108.}

\bibitem{zhu2019transferable}
\href{http://proceedings.mlr.press/v97/zhu19a.html}
{Chen Zhu et al. “Transferable clean-label poisoning
attacks on deep neural nets”. In: \emph{International
conference on machine learning}. PMLR. 2019,
pp. 7614–7623.}

\bibitem{10.1145/3576915.3616661}
\href{https://dl.acm.org/doi/abs/10.1145/3576915.3616661}
{Yi Zhu et al. “TileMask: A Passive-Reflection-based
Attack against mmWave Radar Object Detection in
Autonomous Driving”. In: \emph{Proceedings of the
2023 ACM SIGSAC Conference on Computer and
Communications Security}. CCS ’23. , Copenhagen,
Denmark, Association for Computing Machinery, 2023,
pp. 1317–1331. \textsc{isbn}: 9798400700507.
\textsc{doi}: 10.1145/3576915.3616661.}
\end{thebibliography}

\newpage
\begin{wrapfigure}{l}{25mm}\includegraphics[width=1in,height=1.25in,clip,keepaspectratio]{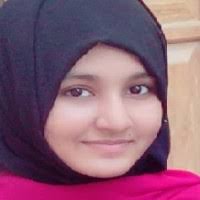}\end{wrapfigure} \par \textbf{Anum Paracha} is a PhD student at the School of Computing and Digital Technology, Birmingham City University, UK. Her research interests are to investigate use of advanced machine learning techniques to mitigate emerging cybersecurity research challenges.
\newline
\begin{wrapfigure}{l}{25mm}\includegraphics[width=1in,height=1.25in,clip,keepaspectratio]{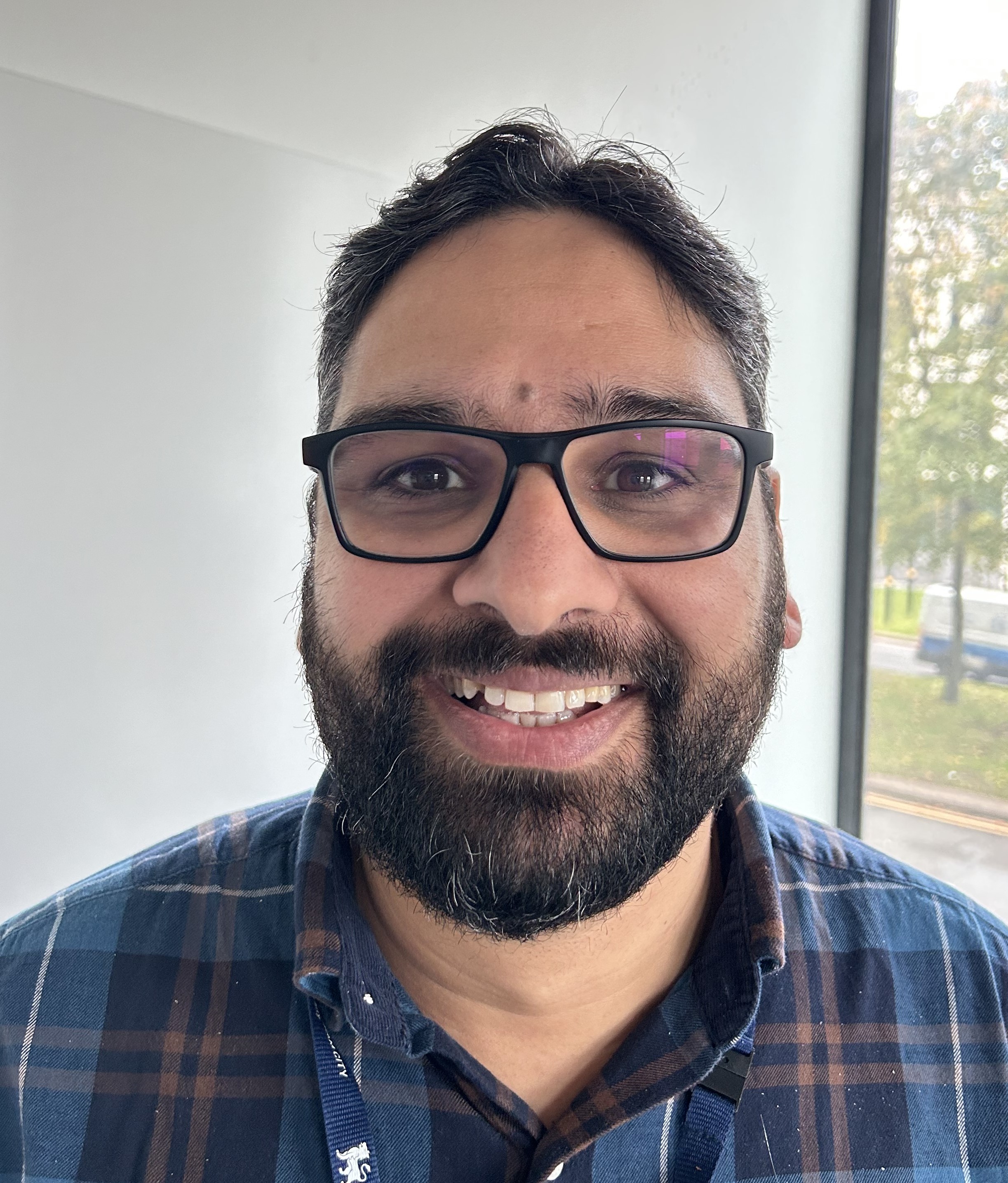}
\end{wrapfigure}\par \textbf{Junaid Arshad} is a Professor in Cyber Security and has extensive research experience and expertise in investigating and addressing cybersecurity challenges for diverse computing paradigms. Junaid has strong experience of developing bespoke digital solutions to meet industry needs. He has extensive experience of applying machine learning and AI algorithms to develop bespoke models to address specific requirements. He is also actively involved in R\&D for secure and trustworthy AI, focusing on practical adversarial attempts on such systems especially as a consequence of cutting-edge applications of generative AI.\par
\bigskip
\begin{wrapfigure}{l}{25mm}\includegraphics[width=1in,height=1.25in,clip,keepaspectratio]{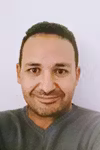}
  \end{wrapfigure}\par\textbf{Mohamed Ben Farah} is a Lecturer in Cyber Security at Birmingham City University. Mohamed has published over 30 journal and conference papers and has organized conferences and workshops in Cyber Security, Cryptography and Artificial Intelligence. He is a reviewer for world-leading academic conferences and journals and is the Outreach Lead of the Blockchain Group for IEEE UK and Ireland.\par
\bigskip
\begin{wrapfigure}{l}{25mm}\includegraphics[width=1in,height=1.25in,clip,keepaspectratio]{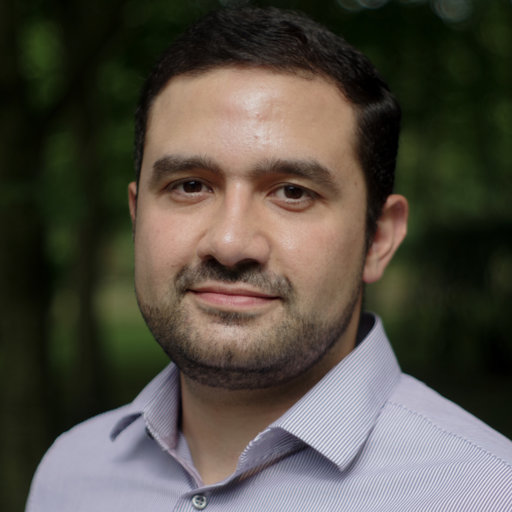}
\end{wrapfigure}\par\textbf{Khalid Ismail} is a Senior Lecturer in Computer Science at Birmingham City University. Dr Ismail’s primary research interests lie in the fields of Artificial Intelligence, computer vision, advanced machine learning, image processing, and deep learning, particularly when applied to complex real-world challenges. Currently, he is supervising many AI based intelligent projects development and also been an active part of industry based collaborative projects.\par
\end{document}